\documentclass{article}

\usepackage{PRIMEarxiv}

\usepackage[utf8]{inputenc} 
\usepackage[T1]{fontenc}    
\usepackage{hyperref}       
\usepackage{url}            
\usepackage{booktabs}       
\usepackage{amsfonts}       
\usepackage{nicefrac}       
\usepackage{amsmath,amssymb}
\usepackage{microtype}      
\usepackage{lipsum}
\usepackage{array}
\usepackage{gensymb}
\usepackage{caption} 
\usepackage{subcaption}
\usepackage{multirow}
\usepackage{fancyhdr}   
\usepackage{graphicx}       
\graphicspath{{Images/}}     

\title{Real-Time Activity Recognition and Intention Recognition Using a Vision-based Embedded System
 
}

\author{
  Sahar Darafsh \\
  Computer Engineering Department \\
  Amirkabir University of Technology\\
  Tehran, Iran \\
  \texttt{sdrfsh@aut.ac.ir} \\
   \And
  Saeed Shiry Ghidary \\
  Mathematics and Computer Science Department \\
  Amirkabir University of Technology \\
  Tehran, Iran \\
  \texttt{shiry@aut.ac.ir} \\
  \And
  Morteza Saheb Zamani \\
  Computer Engineering Department \\
  Amirkabir University of Technology \\
  Tehran, Iran \\
  \texttt{szamani@aut.ac.ir} \\
 }

\begin{document}
\maketitle

\begin{abstract}
With the rapid increase in digital technologies, most fields of study include recognition of human activity and intention recognition, which are essential in smart environments. In this study, we equipped the activity recognition system with the ability to recognize intentions by affecting the pace of movement of individuals in the representation of images. Using this technology in various environments such as elevators and automatic doors will lead to identifying those who intend to pass the automatic door from those who are passing by. This system, if applied in elevators and automatic doors, will save energy and increase efficiency. For this study, data preparation is applied to combine the spatial and temporal features with the help of digital image processing principles. Nevertheless, unlike previous studies, only one AlexNet neural network is used instead of two-stream convolutional neural networks. Our embedded system was implemented with an accuracy of 98.78\% on our intention recognition dataset. We also examined our data representation approach on other datasets, including HMDB-51, KTH, and Weizmann, and obtained accuracy of 78.48\%, 97.95\%, and 100\%, respectively. The image recognition and neural network models were simulated and implemented using Xilinx simulators for the Xilinx ZCU102 board. The operating frequency of this embedded system is 333 MHz, and it works in real-time with 120 frames per second (fps).
\end{abstract}


\section{Introduction}
With the exponential rise in digital technologies, there is a need to implement an intelligent machine to detect human actions and recognize them \cite{SanalKumar2020}. Activity recognition has a variety of uses in various areas, such as surveillance for abnormal or criminal activities, monitoring older people's activities, individual assistants, human-computer interactions \cite{Bux2017}, personal biometric signatures \cite{Sunny2015}, or video retrieval by activity category \cite{Amirbandi2016}. Human Activity Recognition (HAR) systems aim to immediately recognize what action is done in the environment \cite{Amirbandi2016}. Several small non-visual sensors have been integrated into wearable devices in recent years, making them useful for human activity detection and recognition in real-world applications \cite{Sunny2015}. In contrast, cameras are deployed everywhere and can be used in vision-based HAR. The primary differentiation between vision-based HAR and non-visual sensor-based HAR is the way the data is interpreted. Visual sensors provide data in the form of images or videos in 2D or 3D, while other sensors provide data in a one-dimensional signal format \cite{Bux2017}. However, non-visual sensors have some difficulties in real applications, such as the need to be worn and run continuously and their battery life \cite{Sunny2015} and on the other hand, vision-based HAR has various difficulties, such as variations in the shape of the body, pose, and movement velocities, diversity in illuminations, angle, and location of the camera \cite{Amirbandi2016}.
Furthermore, in some activity detection systems, the speed at which people move influences the system's response. For instance, in situations where a person runs to the elevator, intends to use the elevator while the door is closing. So if the elevator is equipped with an activity recognition system, the door can remain open for a longer time. However, if someone is walking around the elevator, he may not have the intention to use the elevator.
In this paper, an image processing system is applied to overcome the limitations of variations in body shape, speed of movement, and camera location. It is then used in the development of an embedded real-time system.

In summary, we make the following contributions:
\begin{itemize}
\item We propose a novel approach to video representation that:
\begin{itemize}
\item It does not require RGB information, 
\item It is able to use the velocity attribute in body movement, 
\item It is not related to camera angle,
\item It can perform even when the scene is crowded.
\end{itemize}
\item Unlike previous researches, we use only one Convolutional Neural Network (CNN) instead of two-stream CNNs to classify video data.
\item We evaluate our image processing method on our intention recognition dataset, HMDB-51, KTH, and Weizmann, and despite the difference in the number of people in the scene, camera angle, and lighting conditions, we achieved high accuracy in each dataset.
\item Our image processing method and classification are performed with high precision and are appropriate for real-time, low-cost embedded systems.
\end{itemize}
The rest of the paper is arranged as follows: Section \ref{sec:related} reviews the most recent works. Section \ref{sec:proposed} describes our novel method for representing videos and classification used for a vision-based intention recognition real-time embedded system. Then, in Section \ref{sec:experiments}, we discuss our experiment and results in different datasets. Finally, conclusions are presented in Section \ref{sec:conclusion}.

\section{Related Works}
\label{sec:related}
\begin{figure}
  \includegraphics[width=\linewidth]{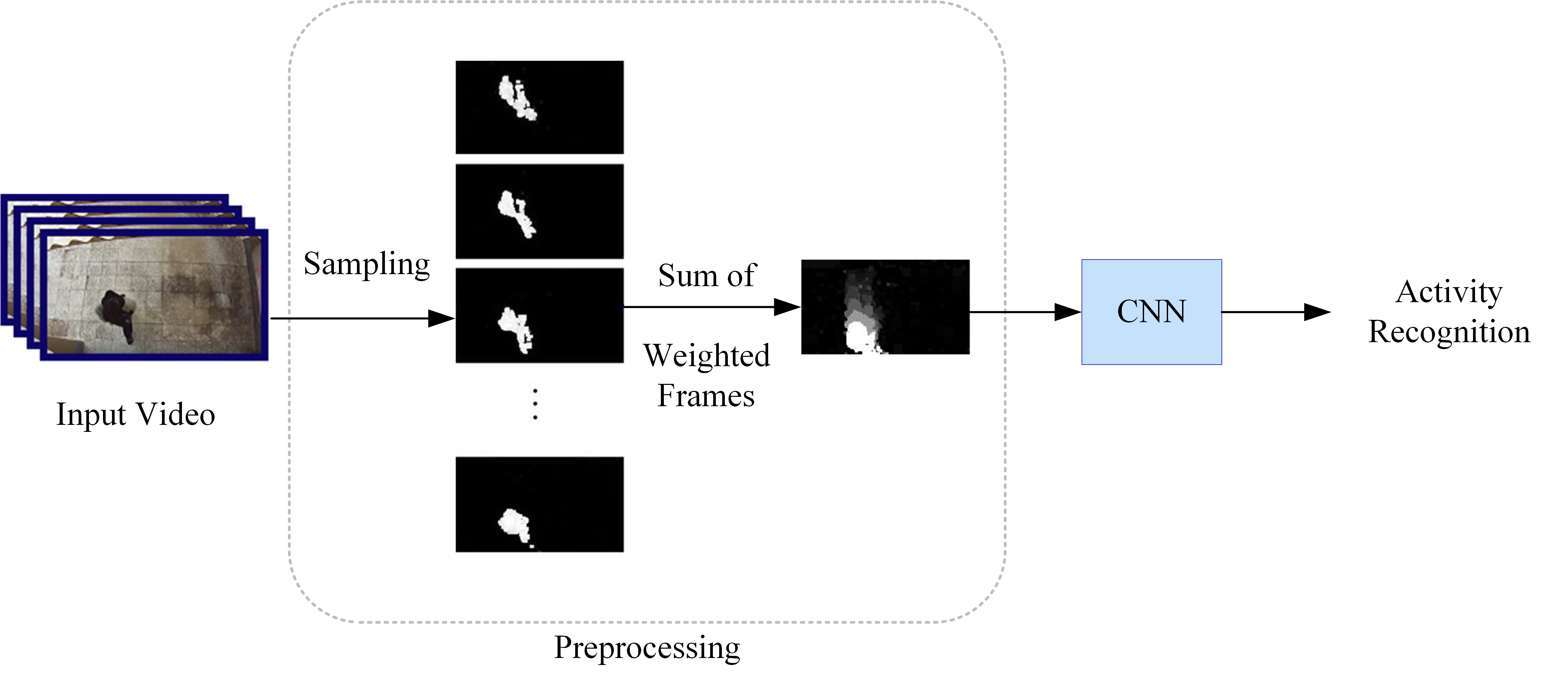}
  \caption{Overview of our approach for activity recognition}
  \label{fig:overview}
\end{figure}

Over the last few years, various algorithms have been developed for human action recognition that have achieved high accuracy. In addition, researchers have introduced handcrafted and Deep Neural Network (DNN) based approaches to action recognition over the last decade \cite{Bux2017}.
Fig. \ref{fig:overview} shows the general structure of a vision-based HAR system. The inputs can be either an image or a sequence of images (video). Preprocessing is performed in the second stage to eliminate the noise and extract features. The extraction of motion features of moving objects in the video is performed in the third stage, and finally, the type of activity is identified \cite{Bodor2003}.

\subsection{Handcrafted Approach}
Earlier works were based on handcrafted features. Such methods extract low-level features from the data and then feed them into a classifier. Handcrafted feature extraction methods' performance relies highly on the feature extraction technique \cite{Chen2020,Huynh2005}. In the Patil et al. method \cite{Patil2018}, after reducing noise and removing the background with the Gaussian Mixture Model (GMM), the Histogram of Oriented Gradient (HOG) was used to extract the image features, the extracted features were then classified with the Support Vector Machine (SVM). Qazi et al. \cite{Qazi2018} obtained the feature vector by combining two HOG features and a Scale Independent Feature Transform (SIFT) and used an SVM to categorize activities. In another study, Chou et al. \cite{Chou2018} proposed a multi-view activity recognition system, using Prewitt and Gabor filters for feature extraction. They used Gaussian Mixture Model Classifier (GMMC), Nearest Neighbor Classifier (NNC), and Nearest Mean Classifier (NMC) for data classification. Vishwakarma \cite{Vishwakarma2020} used activity recognition by Gabor Wavelet Transform (GWT), Ridgelet Transform (RT), and optical flow for feature extraction, and then they performed the classification with K-Nearest Neighbors (KNN). Singh and Vishwakarma \cite{Singh2019} investigated the recognition of activity in low-quality videos using human body silhouettes and SVM.

\subsection{Deep Learning Approach}
Recently, deep learning has been successfully employed for visual categorization tasks \cite{Bux2017}. Recent studies have shown no method of manually selecting attributes that work optimally on all datasets, so extracting attributes from raw data would be more efficient. Ko and Sim \cite{Ko2018} conducted a research study that aims to detect abnormal activities in real-time. In this study, the main subjects of the image are extracted in real-time using YOLOv2 so that all subjects are labeled the same. The results are then fed into a CNN network integrated with a Kalman filter, which distinguishes the subjects. Finally, it is given to another CNN network, followed by an LSTM to detect abnormal activity. Asghari-Esfeden et al. \cite{Asghari-Esfeden2020} introduced the DynaMotion method to represent video data information and used the appearance and motion features together. They used the pose estimation block to extract the characteristics of the joint position, the heat map of each frame, and an encoder to keep the data representation size constant. Finally, they classified the data using CNN and evaluated the results of their research on various datasets. In another research, Abdelbaky et al. \cite{Abdelbaky2020} introduced the Principal Component Analysis Network (PCANet) to minimize computations, considering the problem of activity detection in 3D space-time needs more complicated computations in DNN. Long Short-Term Memory (LSTM) neural networks and CNNs have achieved excellent results in activity detection so far. Deep learning methods can process raw audio or video data and automatically extract features \cite{Chen2020,Huynh2005}. As the human brain uses separate pathways to recognize objects and motions, two-stream neural networks were presented by Simonyan and Zisserman \cite{Simonyan2014} to classify activities in videos. They used a CNN stream for spatial features and another CNN stream for temporal features. The spatial stream performs action recognition from video frames, while the temporal stream is equipped to identify motion movement in dense optical flow. Zhang et al. \cite{Zhang2018} suggested applying motion vectors to prevent high optical flow calculations. However, as motion vectors are extremely noisy and cause neural network training to be inaccurate, they proposed a novel approach of deeply transferred motion vector CNN: the optical flow training phase of the neural network and the motion vector test phase. Finally, the approach performed faster than optical flow. Hou et al. \cite{Hou2017} introduced a Tube Convolution Neural Network (T-CNN) to utilize 3D convolution within the CNN. Researchers have found that LSTM neural networks are extremely accurate for complicated activities because CNNs perform well in short-term, simple tasks but inadequately in long-term and complex ones. Sun et al. \cite{Sun2017} introduced a lattice LSTM network. In this approach, two CNNs are followed by two streams of LSTM neural networks. CNNs are used for extracting the features in both spatial and temporal domains. Spatial network inputs are RGB channels, and the optical flows extracted from the frames are supplied to the temporal neural network as input. One limitation of the LSTM neural network is its redundancy and high computational cost. The difference between this method and the LSTM-based methods is that it has two streams of neural networks, which makes the network, in addition to preserving the time domain features, learn the spatial domain features. 
When raw data is video, the frames are sequentially read and their properties extracted, while not all frames contain new information in the video data. Therefore, extracting the feature from all frames would lead to the extraction of unnecessary information, increase the computation and the training time of the network. Researchers used methods to preprocess videos to solve this problem. Using the concept that a small number of frames in a video provide enough information to identify different activities, Kar et al. \cite{Kar2017} introduced the AdaScan approach for estimating the importance of the frames and then used the most significant frames for the deep neural network. Korbar et al. \cite{Korbar2019}) used another method by applying sampling in long videos, in which shorter samples are selected from long videos, grouped into similar categories, and scored. To minimize computational costs, they used shorter clips with higher scores in the deep learning framework. Ullah et al. \cite{Ullah2017} selected the sixth one for every six frames of video data to minimize redundancy and increase the performance of the LSTM neural network model and extracted its features using CNN. Then, the frames are inserted into a deep bidirectional LSTM neural network during the training phase. The bi-directional LSTM neural network improves the model's computing power and allows for sequential data processing. In this network, the initialization of the neural network is very important because of the huge quantity of calculations. This process requires measuring the person's exact location in the scene and changes in the person's body and the movement of objects. As a result, it will not be suitable for crowded scenarios. It is also obvious that unneeded computations are frequently performed as part of the calculation process, which increases the computation cost \cite{Sun2017}. Finally, Bilen et al. \cite{Bilen2018} introduced Dynamic Images for video preprocessing. The input images result from optical flow and dynamic optical flow mapped onto the static RGB image, generating Dynamic Images and feeding them into a neural ResNet50 classification network. Activity recognition methods have also been used in research related to intention recognition. Rasouli et al. \cite{Rasouli2019} conducted research to recognize people's intent to cross the street. People's next movement is predicted using their previous trajectories and to identify people who intend to cross the street.

\subsection{Real-Time Embedded Systems}
Embedded systems are designed to perform best for a specific application at the lowest cost, whereas general-purpose computers, such as personal computers, are designed to meet general needs. Embedded systems can be utilized in real-time applications. In addition to selecting a proper processor, the choice of an optimal algorithm also plays a critical role in real-time processing. Nowadays, DNNs are among the latest developing artificial intelligence models contributing to many advances in different industries. Real-time computation is one of the challenges of DNNs in embedded systems. One of the constraints of implementing DNNs on embedded processors is their limited bandwidth and internal memory \cite{Mao2018}. In neural networks, more features are extracted from raw data as the depth of the network increases. As the depth of the neural network grows, more memory and computing power are required for real-time processing. In this case, the main challenge for controlling the depth of the neural network is that the calculation of unnecessary features should be avoided to minimize computation and the use of memory \cite{Suto2019}. Cameron et al. \cite{Cameron2019} investigated different and effective parameters to implement a surveillance embedded system based on CNNs. Cameron et al. compared different CNN architectures on different embedded processors and showed that the AlexNet architecture performed better on these processors than on other architectures.
Application-Specific Integrated Circuits (ASICs) are integrated circuits designed and optimized for specific operations. ASIC has recently been used to implement CNNs. ASIC design is very efficient in terms of power consumption, but these devices have a fixed and unchangeable structure and incur high manufacturing costs \cite{Desoli2017}.
Monteiro et al. \cite{Monteiro2018} used Raspberry Pi to implement their embedded system. This device aims to detect objects by processing and to track the motion of the image. They used Local Binary Pattern (LBP) to extract image features, which performs very quickly in the classifier's training.
FPGA is another hardware platform that is appropriate for implementing embedded systems in terms of power consumption and implementation costs. Researchers have used solutions such as data compression \cite{Guo2018,Wu2018}, binary inputs and calculations \cite{Liang2018,Nakahara2017}, and computational parallelization \cite{Ignatov2018,Ma2018} to speed up the system's response time by implementing CNNs on FPGAs. In another approach, Liang et al. \cite{Liang2020} used the Fast Fourier Transform (FFT) to perform convolutional calculations on FPGAs.
In embedded systems, the choice of processor is based on the system requirements. Our underlying system in this research is an image processing and categorization system that must operate in real-time.

Implementing the processing part in embedded systems is possible in two ways:
\begin{enumerate}
\item The Programmable Logic (PL) component implemented on the logic blocks and DSP blocks of the FPGA, and
\item The Processor Subsystem (PS) running on a processor such as the ARM processor.
\end{enumerate}
Using an ARM processor causes a significant delay in system processing, but it is simpler to implement image processing algorithms. FPGA is much less delayed and it is more complex to implement such algorithms. By introducing boards for embedded systems, including the PS and PL segments, FPGA design tools have made it possible for designers to take advantage of both methods on a single board. In these boards, the PL segment and the ARM processor are connected by an AXI bus. As a result, certain parts of the system can be implemented as logic circuits in the PL segment and other components can run as PS applications \cite{Ramagond2018}. Using PL and PS parts, Ali Altuncu et al. \cite{AliAltuncu2015} implemented a real-time image processing system on Xilinx boards.

\section{Proposed Approach}
\label{sec:proposed}
\begin{figure}
  \includegraphics[width=\linewidth]{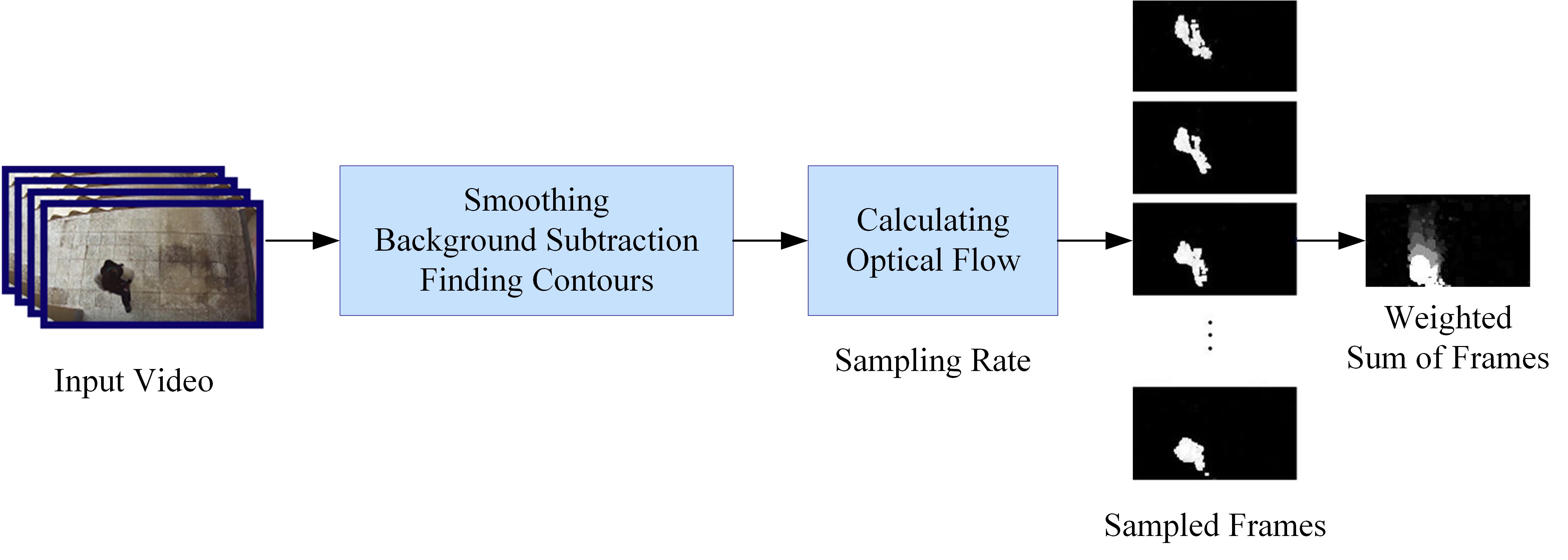}
  \caption{Preprocessing steps}
  \label{fig:preprocessing}
\end{figure}

This section introduces our novel method of representing video data and explains how to add the feature of speed and independence of this representation of the visual details and the camera angle. The main steps for preparing the data for classification are shown in Fig. \ref{fig:preprocessing}.

\subsection{Preprocessing and Tracking}
Preprocessing of the data is performed based on the purpose and amount of information required to reduce unnecessary computations to prepare the data for modeling. In preprocessing, dilation and erosion operations and a proper Gaussian filter are used to eliminate noises. The next step after smoothing the image is to remove the image background. Choosing a reliable algorithm to remove the background is very challenging and error-prone. Background removal challenges include slow transition in image brightness, a sudden change in image illumination, shadows, and dynamic backgrounds \cite{Brutzer2011}. By comparing different methods for removing backgrounds, it can be seen that removing backgrounds with KNN and removing image shadows for real-world data work better under a variety of lighting conditions than other methods \cite{Shallari2018}. We generate a binary video for each data, in which the video's background is black, and the moving subject is white. The condition for the presence of a moving contour in the image is considered to be the starting point for tracking \cite{Babiker2018}.
\subsection{Sampling}
Video images are divided into two categories of time and space features. In previous studies, activities were described by calculating the optical flow of temporal features of activities and extracting spatial features from static image frames, where each became part of the CNN stream \cite{Li2018}. This study applies the concept used to generate dynamic images in the Bilen et al. method \cite{Bilen2018} and the average velocity formula in physics. Eq. (\ref{eq:eq1}), $\Delta{x}$ denotes the amount of displacement, while $\Delta{t}$ denotes the time it takes for the displacement to occur.

\begin{equation}
\label{eq:eq1}
\Bar{v} = \frac{\Delta{x}}{\Delta{t}}
\end{equation}

If the sampling rate is a fixed number at the sampling stage, the data in which the individual moves faster than the sampling rate will be missed, and the data in which the individual moves slower than the sampling rate will be redundant. After the background of the videos was removed, the frames were sampled to reduce redundancy and complexity. Consequently, we defined the adaptive sampling rate, which is compatible with people's speed of movement. Since the fps rate is constant for our camera, and all data is received simultaneously, moving between two fixed frames means a fixed time for all data, and moving can be proportional to its speed. We used optical flow to calculate the displacement by calculating magnitudes (distances) and angles (directions). We averaged the two vectors separately, and by doing so, we ensure that the average displacement is extracted. Finally, we converted the mean magnitude and mean angle defined in polar coordinates to the Cartesian coordinate system. Ultimately, the adaptive sampling rate is defined as Eq. (\ref{eq:eq2}).

\begin{equation}
\label{eq:eq2}
S = \sqrt{\Bar{u}^2+\Bar{v}^2}
\end{equation}

\begin{figure*}[t!]
    \centering
    \begin{subfigure}[t]{0.3\textwidth}
        \centering
        \includegraphics[height=1.2in]{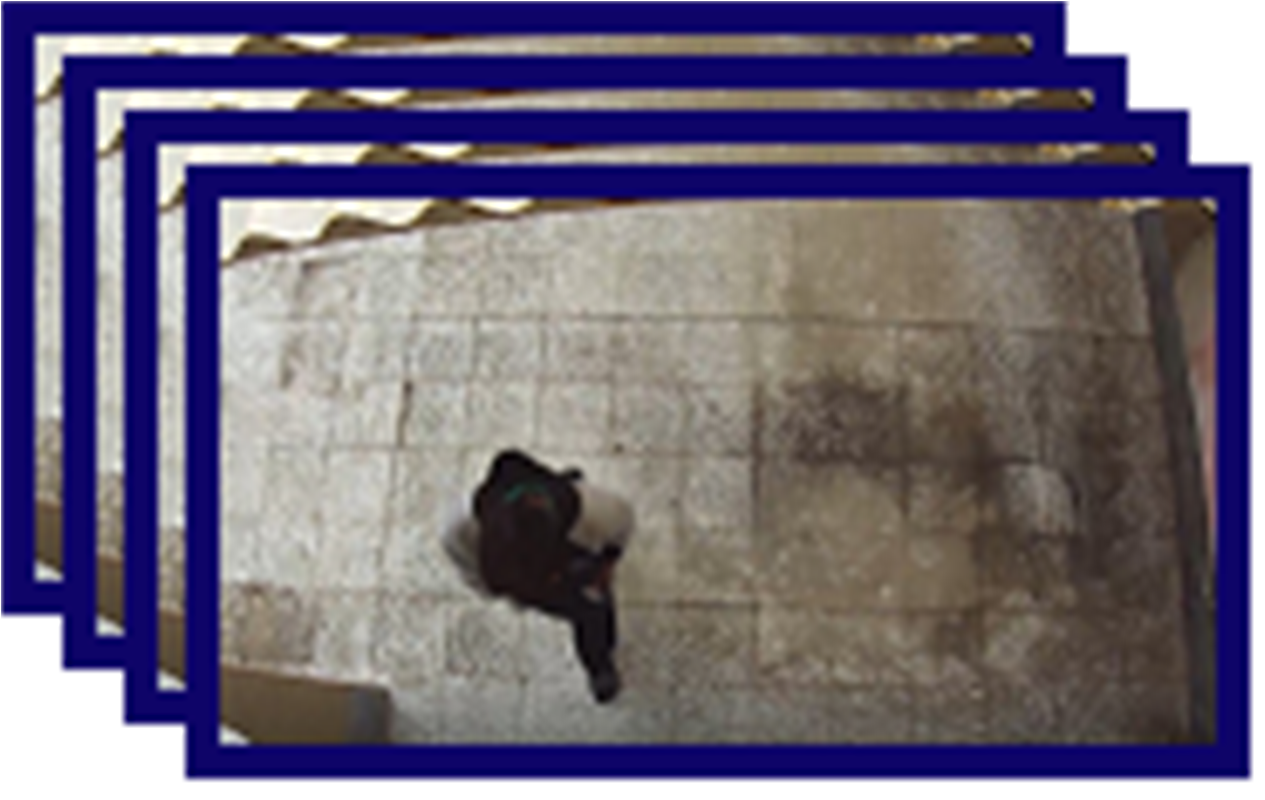}
        \caption{}
    \end{subfigure}%
    ~ 
    \begin{subfigure}[t]{0.34\textwidth}
        \centering
        \includegraphics[height=1.2in]{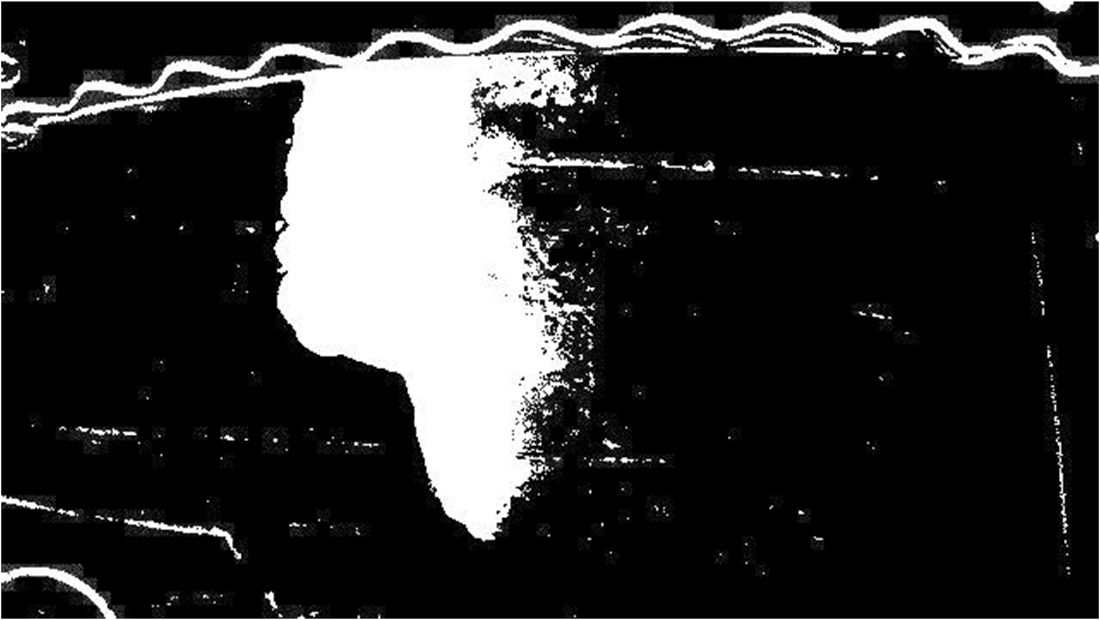}
        \caption{}
    \end{subfigure}%
    ~
    \begin{subfigure}[t]{0.34\textwidth}
        \centering
        \includegraphics[height=1.2in]{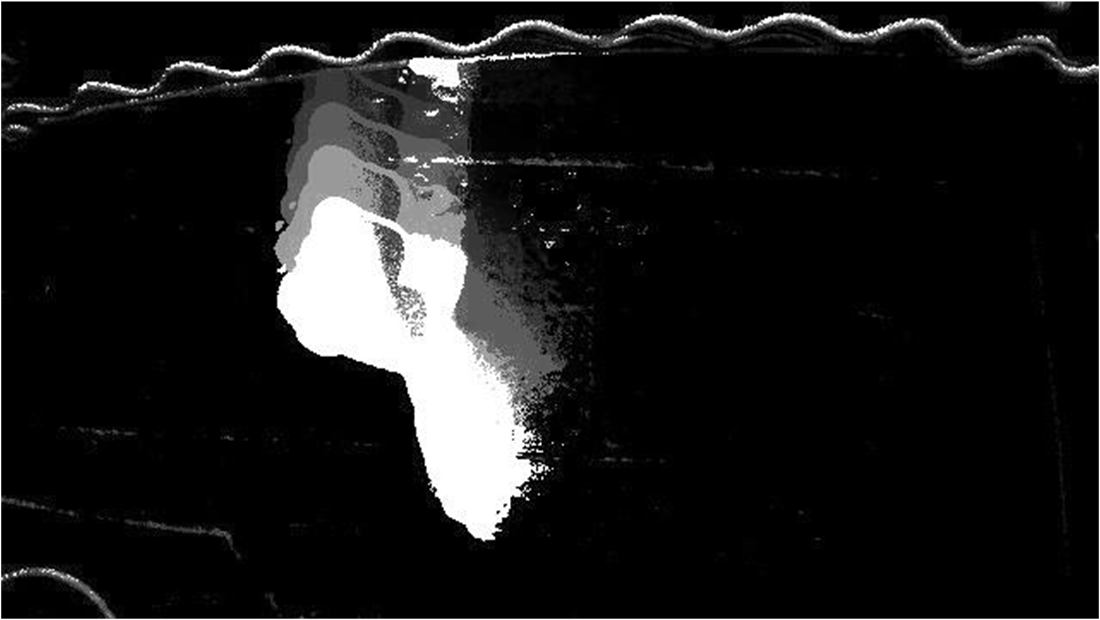}
        \caption{}
    \end{subfigure}
    \caption{(a) Original input, (b) The sum of non-weighted frames, (c) The sum of weighted frames}
    \label{fig:weighted_nonweighted}
\end{figure*}

In Eq. (\ref{eq:eq2}), $S$ is the sampling rate, $\Bar{u}$ is the mean value of the amplitude and $\Bar{v}$ is the mean value of the angles. After sampling, we removed five samples from the beginning and five samples from the end of each training data sample to reduce the training process's errors. It allows the model to concentrate on the individual's effective behavior. Because the system must respond to the most recent behavior of individuals, the most recent sampled frames are more significant in this representation. Therefore, as we add frames to each other, the frames are weighted, and the new frame's addition is done with $\beta$ percent of the previous frames. This weighting continues from the first frame to the last, and each time the previous frames are decreased by $(1-\beta)$ of their intensity, the first frames will have further intensity reductions. 
Fig. \ref{fig:weighted_nonweighted} shows the difference between the sum of the weighted and non-weighted frames. One problem is dealing with unwanted noise, leading to zero or infinite values for the sampling rate in some cases. We set the sampling rate to a low value to fix this concern, such as two, and avoid data loss in exchange for redundancy for zero and infinite values.

\subsection{Network Architecture and Training}
\begin{figure*}[t!]
    \centering
    \begin{subfigure}[t]{0.3\textwidth}
        \centering
        \includegraphics[height=1.2in]{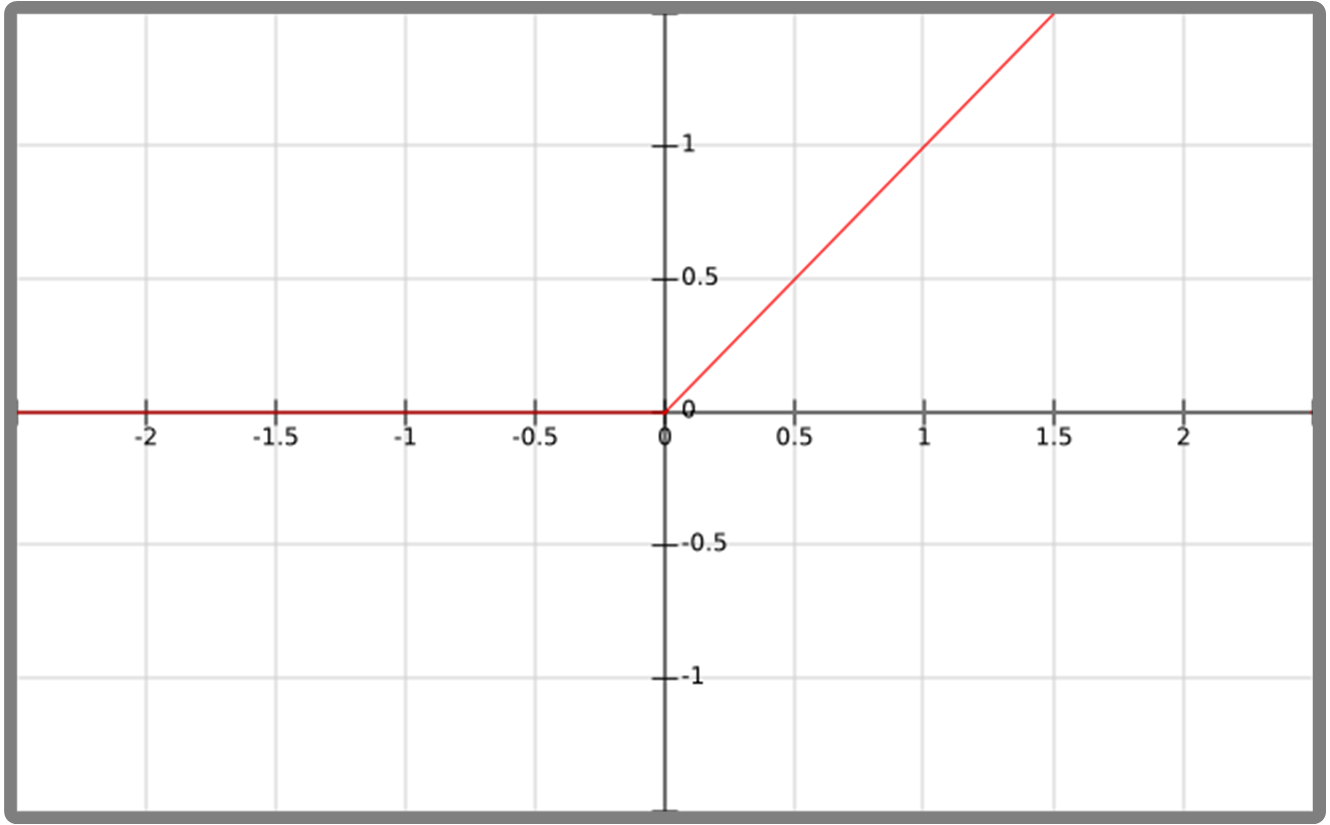}
        \caption{}
    \end{subfigure}%
    ~ 
    \begin{subfigure}[t]{0.3\textwidth}
        \centering
        \includegraphics[height=1.2in]{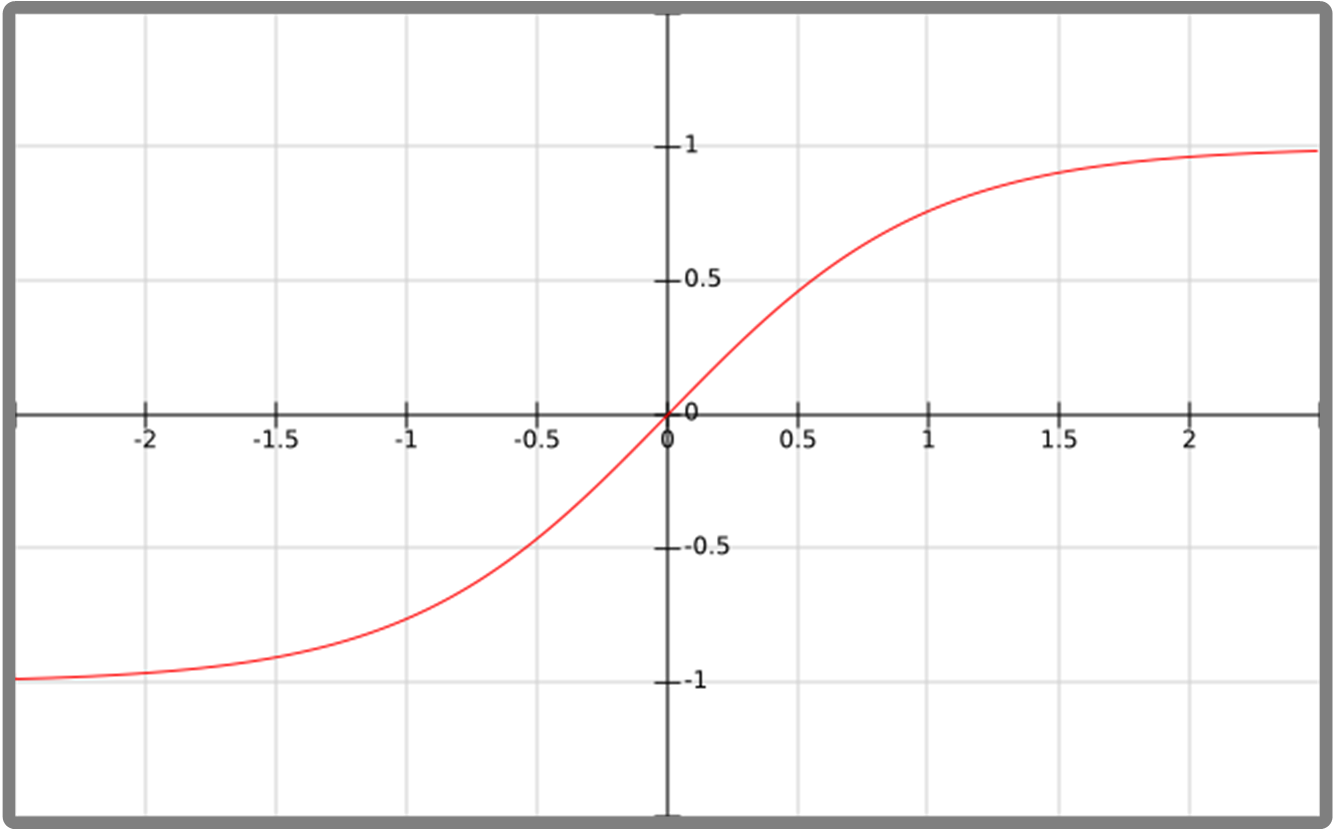}
        \caption{}
    \end{subfigure}%
    ~
    \begin{subfigure}[t]{0.3\textwidth}
        \centering
        \includegraphics[height=1.2in]{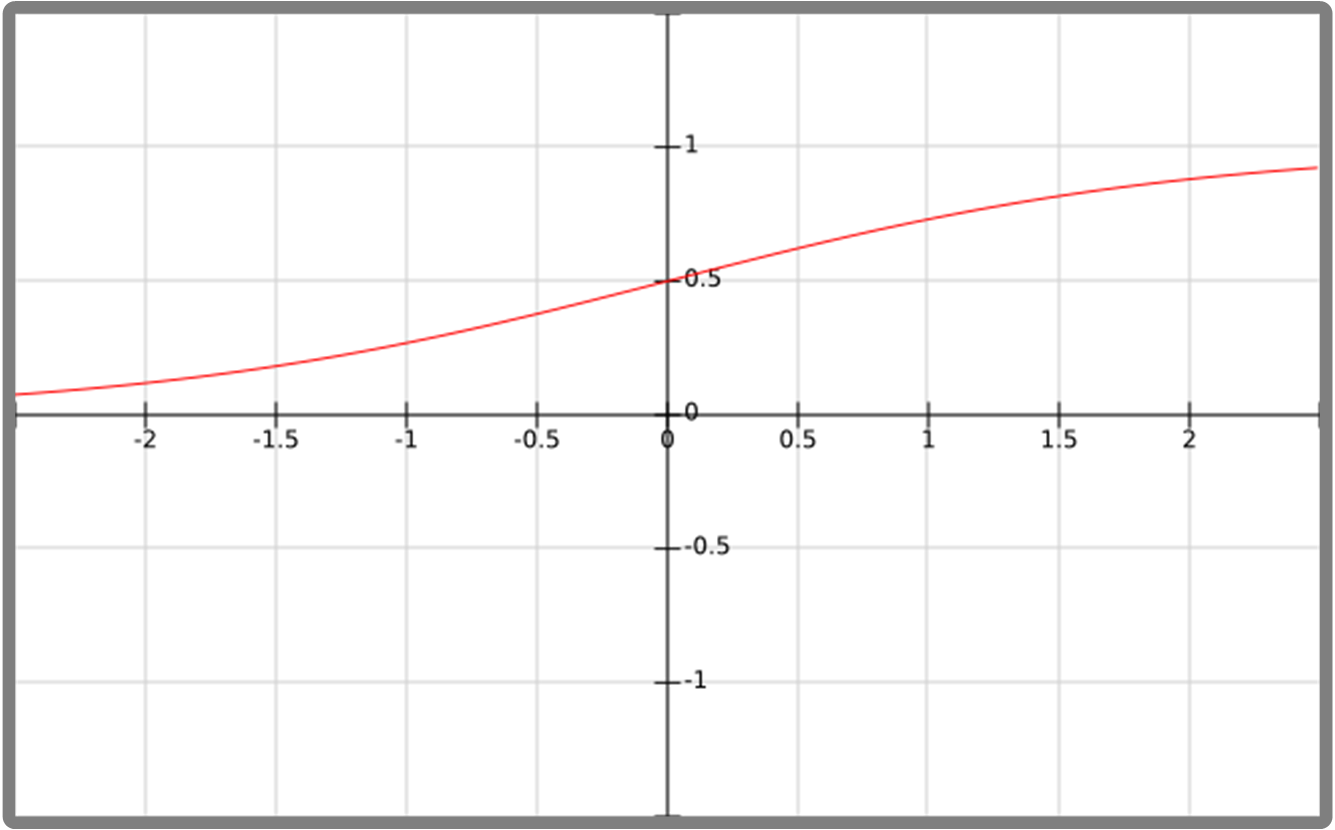}
        \caption{}
    \end{subfigure}
    \caption{(a) ReLU Function, (b) Sigmoid Function, (c) Tanh Function}
    \label{fig:activation}
\end{figure*}

In this research, our purpose is to recognize activities in which the shape of the body moves in the same way, and the speed of movement is the most influential factor in distinguishing these activities. Since we do not need many details and calculations, and the system's computational cost and response time are essential, categorization can be performed with a CNN. From the analysis of Cameron et al. \cite{Cameron2019} study, AlexNet's architecture has a strong performance on embedded processors, so we have classified our data with AlexNet's CNN. AlexNet was developed by Alex Krizhevsky et al. \cite{Krizhevsky2017}. It has five convolutional layers, three max-pooling layers, two normalization layers, two fully connected layers, and one softmax layer, as shown in Table \ref{table:alexnet}.
\begin{table}[]
    \centering
    \renewcommand{\arraystretch}{1.35}
    \caption{Layers of the AlexNet Architecture}
    \label{table:alexnet}
    \begin{tabular}{  c  c  c  c  } 
    \noalign{\hrule height 1.5pt}
    \textbf{Layer} & \textbf{Type} & \textbf{Activation} & \textbf{Filters}\\ 
    \noalign{\hrule height 1.5pt}
    \multirow{2}{3em}{\textbf{Conv1} \textbf{Pool1}} & \multirow{2}{6em}{Conv2D MaxPooling2D} & \multirow{1}{2em}{Tanh} & \multirow{1}{2em}{96} \\
    \\
    \hline
    \multirow{2}{3em}{\textbf{Conv2} \textbf{Pool2}} & \multirow{2}{6em}{Conv2D MaxPooling2D} & \multirow{1}{2em}{Tanh} & \multirow{1}{2em}{256} \\
    \\
    \hline
    \multirow{2}{3em}{\textbf{Conv3} \textbf{Pool3}} & \multirow{2}{6em}{Conv2D MaxPooling2D} & \multirow{1}{2em}{Tanh} & \multirow{1}{2em}{384}
    \\
    \\
    \hline
    \multirow{2}{3em}{\textbf{Conv4} \textbf{Pool4}} & \multirow{2}{6em}{Conv2D MaxPooling2D} & \multirow{1}{2em}{Tanh} & \multirow{1}{2em}{384} \\
    \\
    \hline
    \multirow{2}{3em}{\textbf{Conv5} \textbf{Pool5}} & \multirow{2}{6em}{Conv2D MaxPooling2D} & \multirow{1}{2em}{Tanh} & \multirow{1}{2em}{256} \\
    \\
    \hline
    \multirow{1}{3em}{\textbf{Flatten}} & \multirow{1}{6em}{Flatten} & & \\
    \multirow{3}{3em}{\textbf{FC1} \textbf{FC2} \textbf{FC3}} & \multirow{3}{6em}{Dense \\Dense \\Dense} & \multirow{3}{2em}{Tanh Tanh Softmax} & \multirow{3}{2em}{4096 4096 2} \\
    \\
    \\
    \noalign{\hrule height 1.5pt}

\end{tabular}
\end{table}
The choice of activation function depends on the application of the neural network. The three sigmoid, ReLU, and Tanh functions, are more common functions for neural networks. The ReLU feature has the fastest output than Tanh and Sigmoid, but it does not work precisely for all network training data. Fig. \ref{fig:activation}(a) shows that the ReLU function diagram has zero output for negative input values, leading to a dying ReLU problem. The sigmoid function is shown in Fig. \ref{fig:activation}(b), which is derivable at all points, and its values are in the range (0, 1). The sigmoid function is not symmetrical around the origin, ensuring that all neurons' output values will be the same. This problem can be solved by using the Tanh function (Fig. \ref{fig:activation}(c)), which has a symmetric graph around the origin and has values in the range (-1,1), leading to a better separation between different categories of input data \cite{Sharma2017}. For the last layer, we used the softmax activation function, which is a combination of several sigmoids, to consider the likelihood of data in different classes. The initial weighting using the Glorot distribution in neural networks works best with the Tanh activation function \cite{Glorot2010}. Therefore, we initialized weights with uniform Glorot distribution and normal Glorot distribution for kernel initialization and obtained smoother results from the normal Glorot distribution.
We used the Sparse Categorical Cross-Entropy (SCCE) function as the loss function, which is defined as:
\begin{equation}
SCCE(y, \hat{y}) = - \sum_{i=1}^{C} y_i \log (\hat{y_i})
\label{eq:eq3}
\end{equation}

In Eq. (\ref{eq:eq3}), we assume $(x,y) \in D$ be an input-output pair from the labeled dataset $D$ containing $C$ categories, where $x$ is the input and $y$ is the actual label for $x$, and $y$ is an integer between $0$ and $C$-1. Let us suppose that our neural network $f$ produces a probability $\Vec{f}(x)=\hat{y} \in [0,1]^c$ (For example, with a softmax), where $\hat{y} \in [0,1]^i$  is the $i^{\text{\tiny th}}$ element of $\hat{y}$. The optimization process is the necessary step to train a neural network model to minimize its error rate. There are various functions for optimization; the criteria for choosing between these functions are the convergence speed and network training power or their generalizability. The appropriate optimizer function is selected depending on the application and the importance of each of these two features. Among the various optimization approaches, if the Stochastic Gradient Descent (SGD) function is fine-tuned precisely, it has the slowest convergence speed and will make the model more generalizable. In this application, because of the wide range of data that the system will receive, a higher training capacity would be more critical than the convergence speed. Therefore, we used the SGD optimization function and enhanced the model's convergence speed by fine-tuning the learning rate and momentum values. Finally, after creating the model, the dataset was divided into training, validation, and testing. Then, AlexNet neural network training was performed.

\subsection{Designing Real-Time Embedded System}
\begin{figure}
  \centering
  \includegraphics[width=\linewidth]{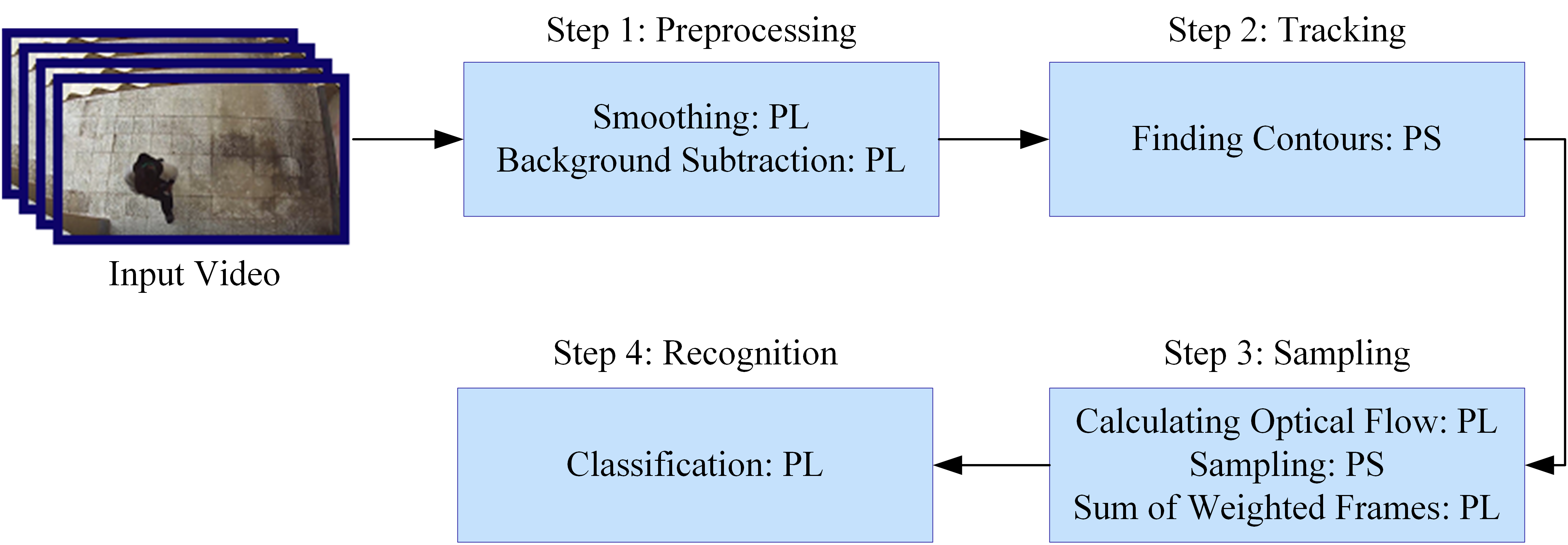}
  \caption{Algorithm steps and the location of each instruction on the ZCU102 board}
  \label{fig:plps}
\end{figure}

The final AlexNet model is frozen and saved in Tensorflow SavedModel file format, including the weights and the computation. In this study, our goal is to simulate a real-time embedded system. We use the PL-PS co-design approach, and considering the high processing capability of the PL part, processes such as neural network calculations and optical flow calculations are delegated to the PL section. We utilized the ZCU102 board containing an FPGA chip from the Zynq UltaScale+ family to simulate the embedded system and take advantage of the ability to process deep neural networks on the PL part; we used DPU IP and synthesized and implemented the system by Vivado 2019.2 software.
Due to the DPU configuration for the ZCU102 board, which contains a ZU9 FPGA device (XCZU9eg-ffv1156-2-e), we used a DPU with $B4096*3$ architecture. Fig. \ref{fig:plps} shows the steps of image processing and the network components and whether these instructions are executed on PL or PS. In Fig. \ref{fig:plps}, the two stages of tracking and sampling are performed periodically in our embedded system. Our system is a periodic embedded system that corresponds to the frames' processing capacity at each step and compares it with the receiving frame rate from the input to prove its promptness. If inputs are received at a slower frame rate than the processing capacity, it can be concluded that the system is in real-time. This will be discussed in section \ref{sec:experiments}.
\\
\\
\section{Experiments and Results}
\label{sec:experiments}
In an overview, as mentioned in section \ref{sec:proposed}, our approach involves calculating the optical flow between two fixed frames. After sampling, the sampled frames are added to each other by weighting them. Sample weighting is done by adding each sample to $\beta$ percent of the previous frame intensity, so the intensity of the older frames decreases by 1-$\beta$ as the new sampled frame is added. In this study, the distance between the two frames used to calculate the optical flow, the intensity of the previous frames, and the size of the optical flow window will affect the final result. In this section, we achieved the best performance of the method on our intention recognition dataset, KTH, Weizmann, and HMDB-51 datasets by examining these modifiable factors in this method, and then we analyze the results of creating an embedded real-time system on the ZCU102 board.
\subsection{Our Intention Recognition Dataset}
\begin{figure*}[t!]
    \centering
    \begin{subfigure}[t]{0.45\textwidth}
        \centering
        \includegraphics[height=1.7in]{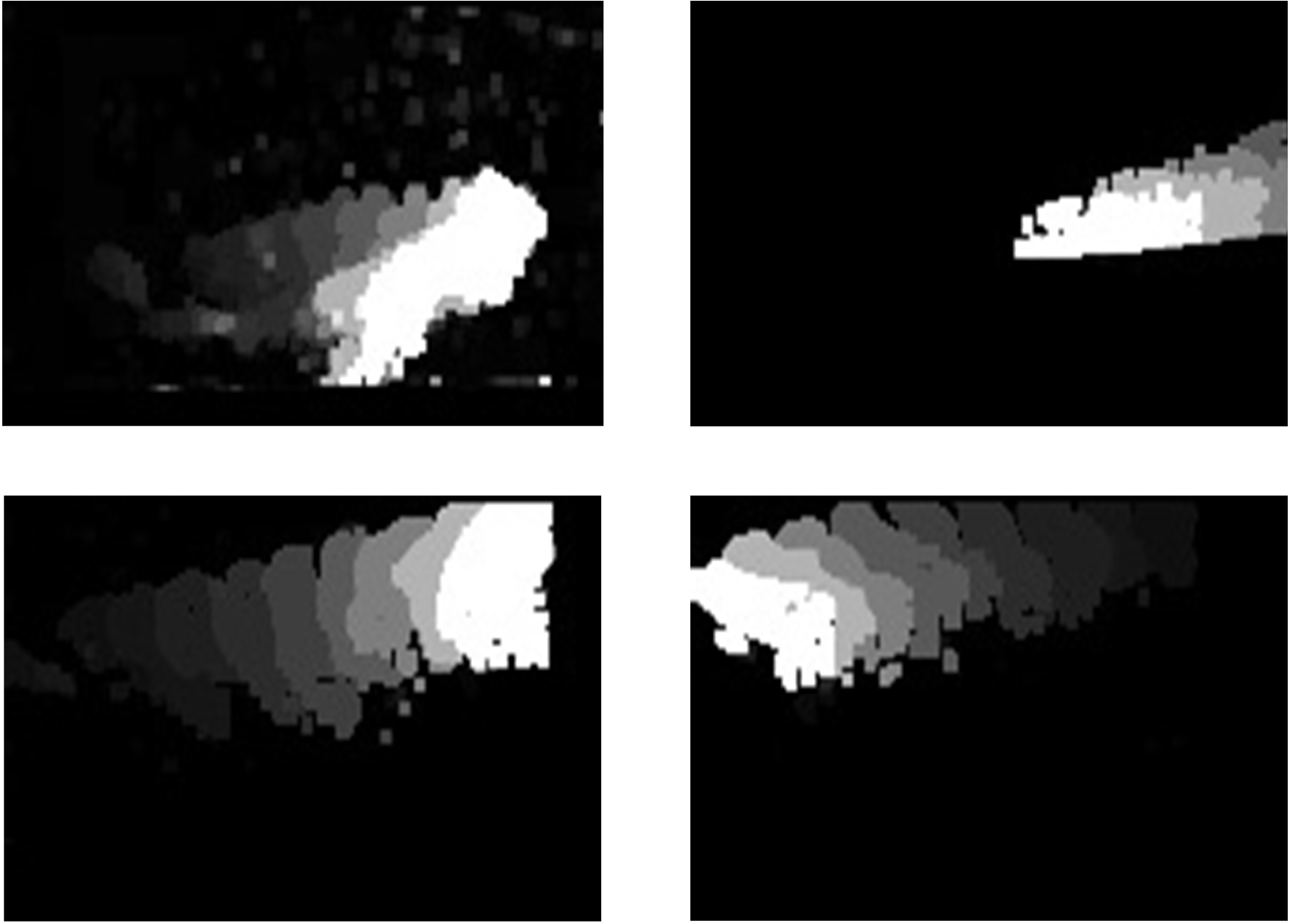}
        \caption{}
    \end{subfigure}%
    ~ 
    \begin{subfigure}[t]{0.5\textwidth}
        \centering
        \includegraphics[height=1.7in]{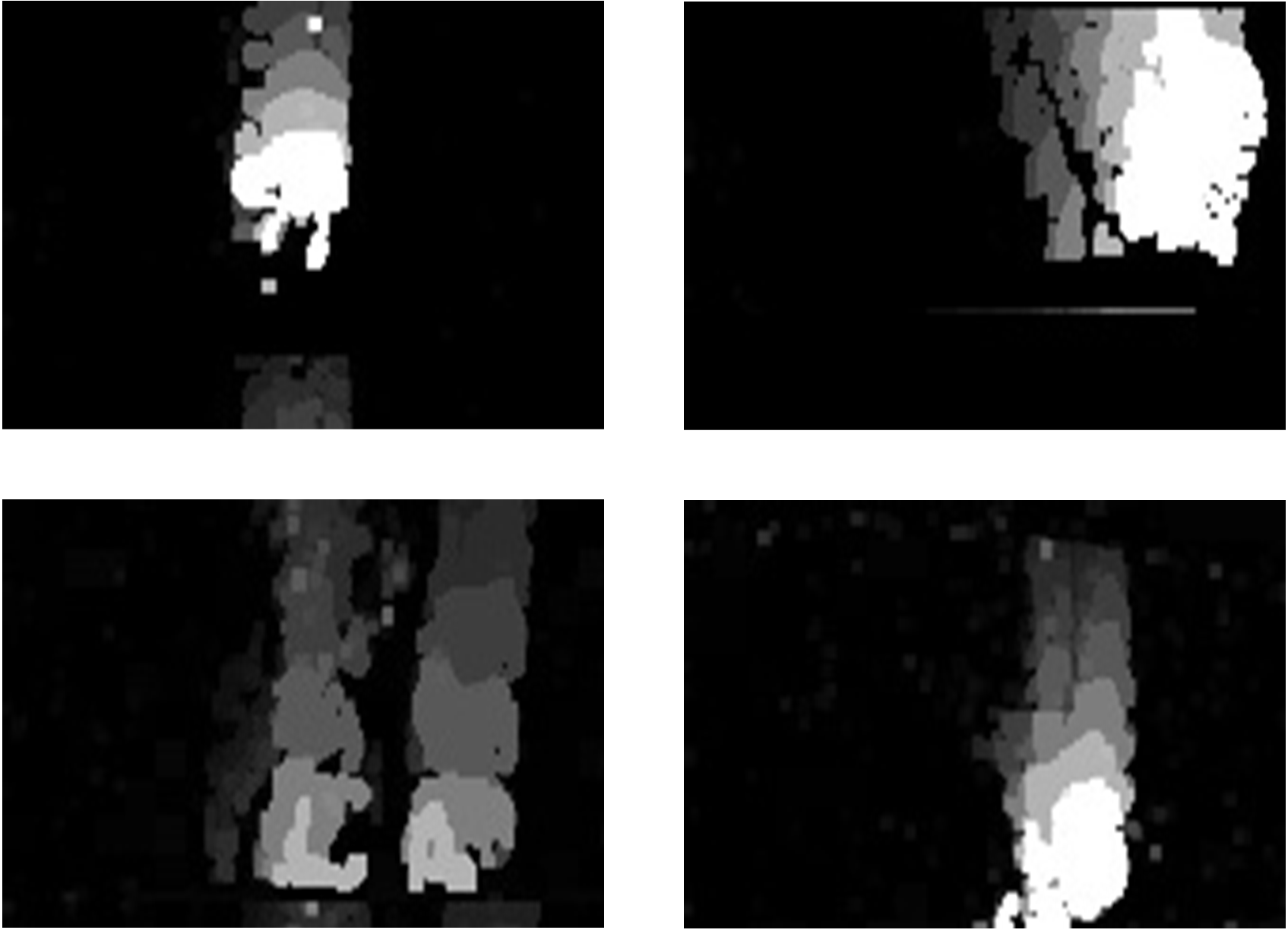}
        \caption{}
    \end{subfigure}%

    \caption{(a) Collected data labeled zero, (b) Collected data labeled one}
    \label{fig:intentiondataset}
\end{figure*}

\begin{figure*}[t!]
    \centering
    \begin{subfigure}[t]{\textwidth}
        \centering
        \includegraphics[height=1.45in]{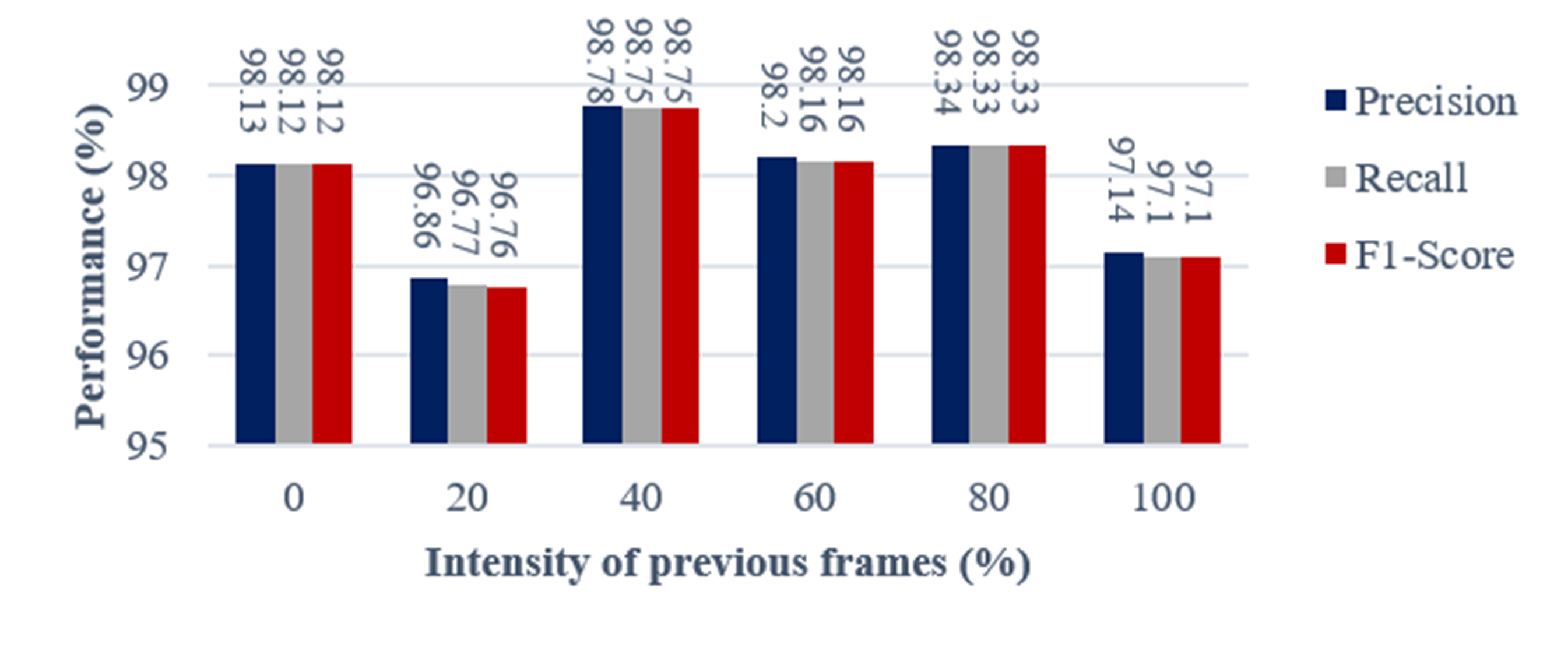}
        \caption{}
    \end{subfigure}%
    \\
    \begin{subfigure}[t]{\textwidth}
        \centering
        \includegraphics[height=1.7in]{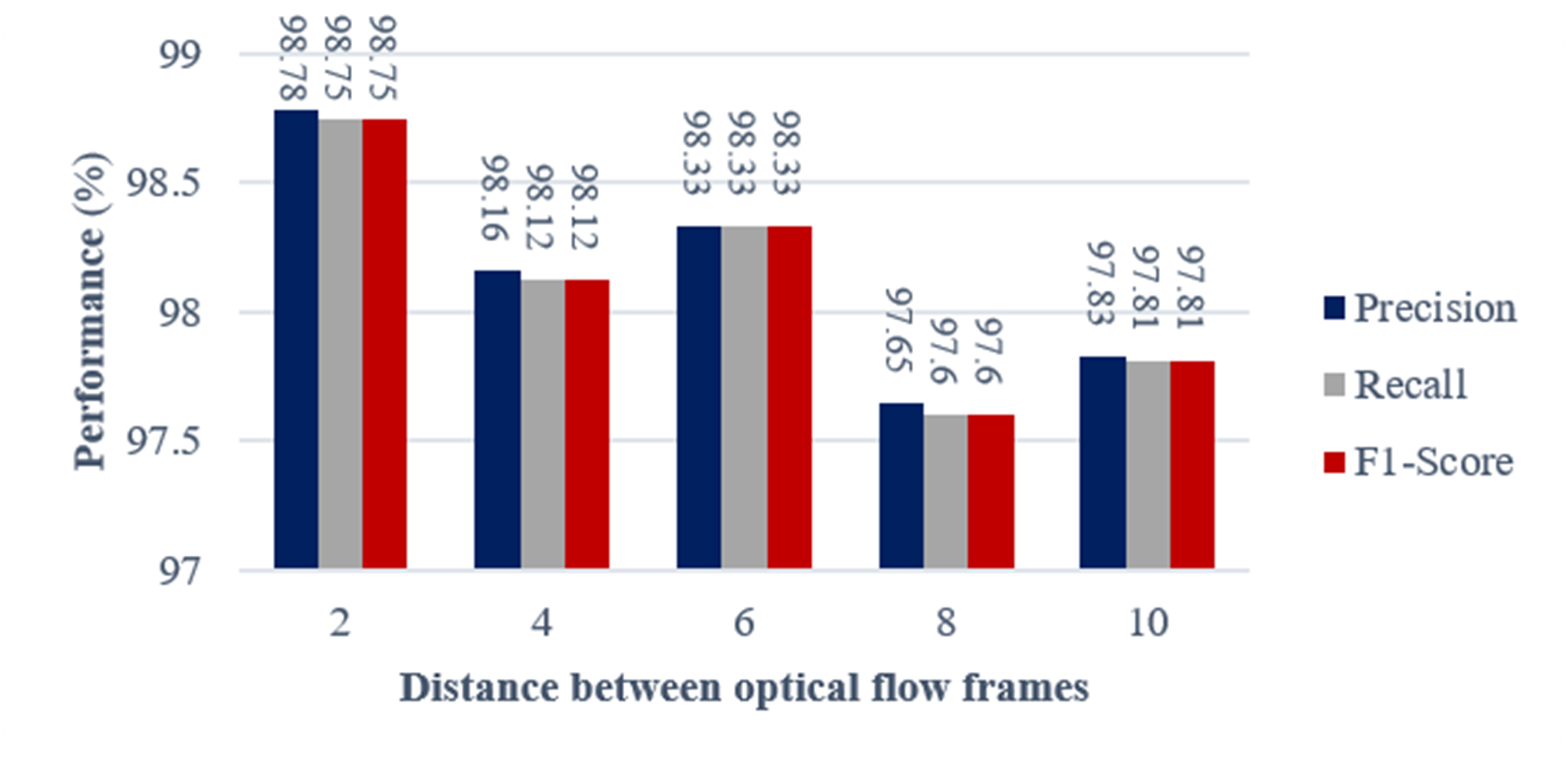}
        \caption{}
    \end{subfigure}%
    \\
    \begin{subfigure}[t]{\textwidth}
        \centering
        \includegraphics[height=1.7in]{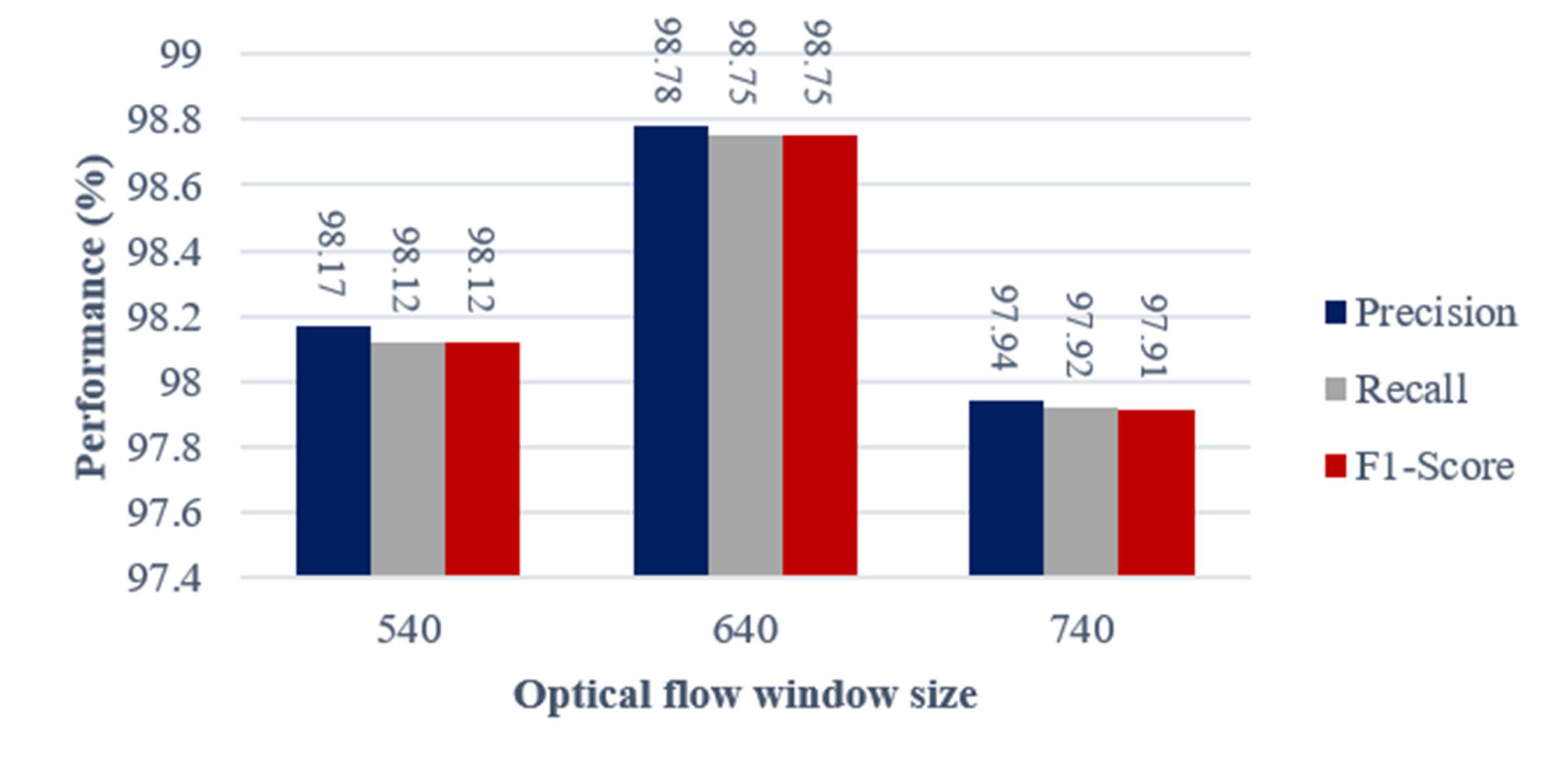}
        \caption{}
    \end{subfigure}
    \caption{(a) The algorithm performance in the Intention Recognition dataset for different values of the previous frames' color intensity, (b) The algorithm performance for calculating optical flow at various distances between frames, (c) The algorithm performance for different optical flow window sizes.}
    \label{fig:intentionexp}
\end{figure*}

The first question in collecting activity recognition data is how many and what activities this model will detect. It is then necessary to consider the details and type of activities, the complexity of the activities, and other scene characteristics \cite{Sigurdsson2017}. Considering given factors, the data were collected in short videos of people by installing a video camera on top of the automatic doors and elevator doors. Several data collection conditions were considered, such as different environments, different movement directions and speeds of people, bicycles and motorcycles crossing the sidewalk, various illumination conditions, and various camera positions.Because this data is collected from the top angle of the head, it does not require much information; nevertheless, when the scene is crowded, one or two individuals from the scene intend to enter the place, while the rest of the individuals in the scene do not intend to enter. This consideration is also provided while collecting data.
Finally, data labeling is performed in two groups: one (for when to trigger the door opening signal) and zero. Approximately 1100 different data are obtained, of which 550 are labeled one and others are labeled zero. Approximately 1100 different data are obtained, of which 550 are labeled one and others are labeled zero.
Fig. \ref{fig:intentiondataset} shows examples of collected data. DNN training depends on the number and variety of training data, and, given that 550 data were collected per class during the data collection process, the network would be overfitted. Therefore, the data augmentation approach has been used to avoid overfitting the neural network. In the data augmentation method, new data is generated by creating random displacements and data mirroring. With the help of this method, the data set of this method has reached 20280 input data and is used for training AlexNet. 
As previously stated, the effective parameters of this approach are the intensity of frames, the distance between optical flow frames, and window size. By modifying these parameters, appropriate data are sampled and generated each time, and their performance results in the AlexNet neural network are evaluated using the parameters precision, recall, and F1-Score. Fig. \ref{fig:intentionexp} demonstrates the experiment results for various values of the color intensity of the frames, the distance between the optical flow frames, and the optical flow window size.

\subsection{KTH Dataset}
\begin{figure*}[t!]
    \centering
    \begin{subfigure}[t]{\textwidth}
        \centering
        \includegraphics[height=1.7in]{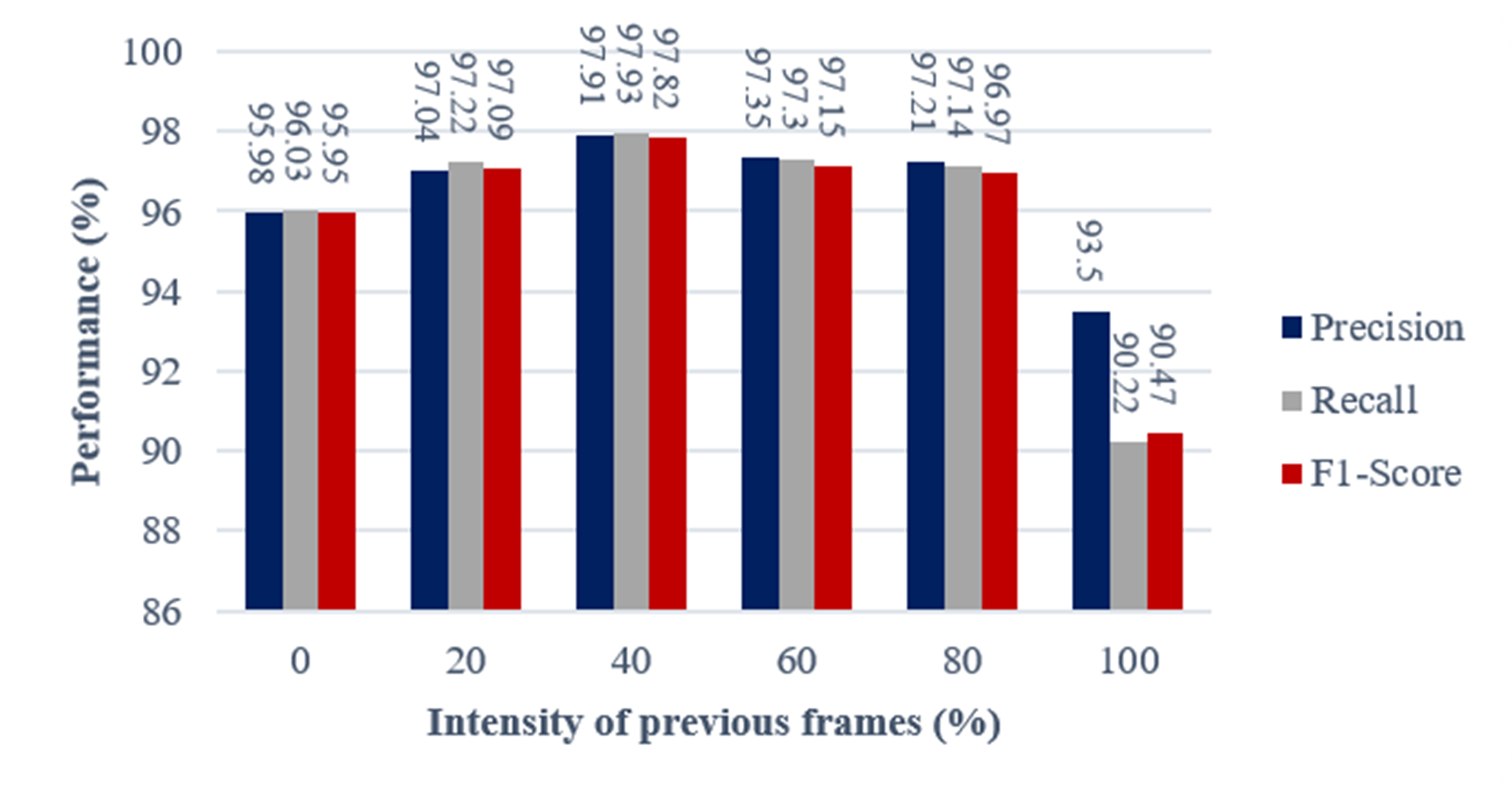}
        \caption{}
    \end{subfigure}%
    \\
    \begin{subfigure}[t]{\textwidth}
        \centering
        \includegraphics[height=1.7in]{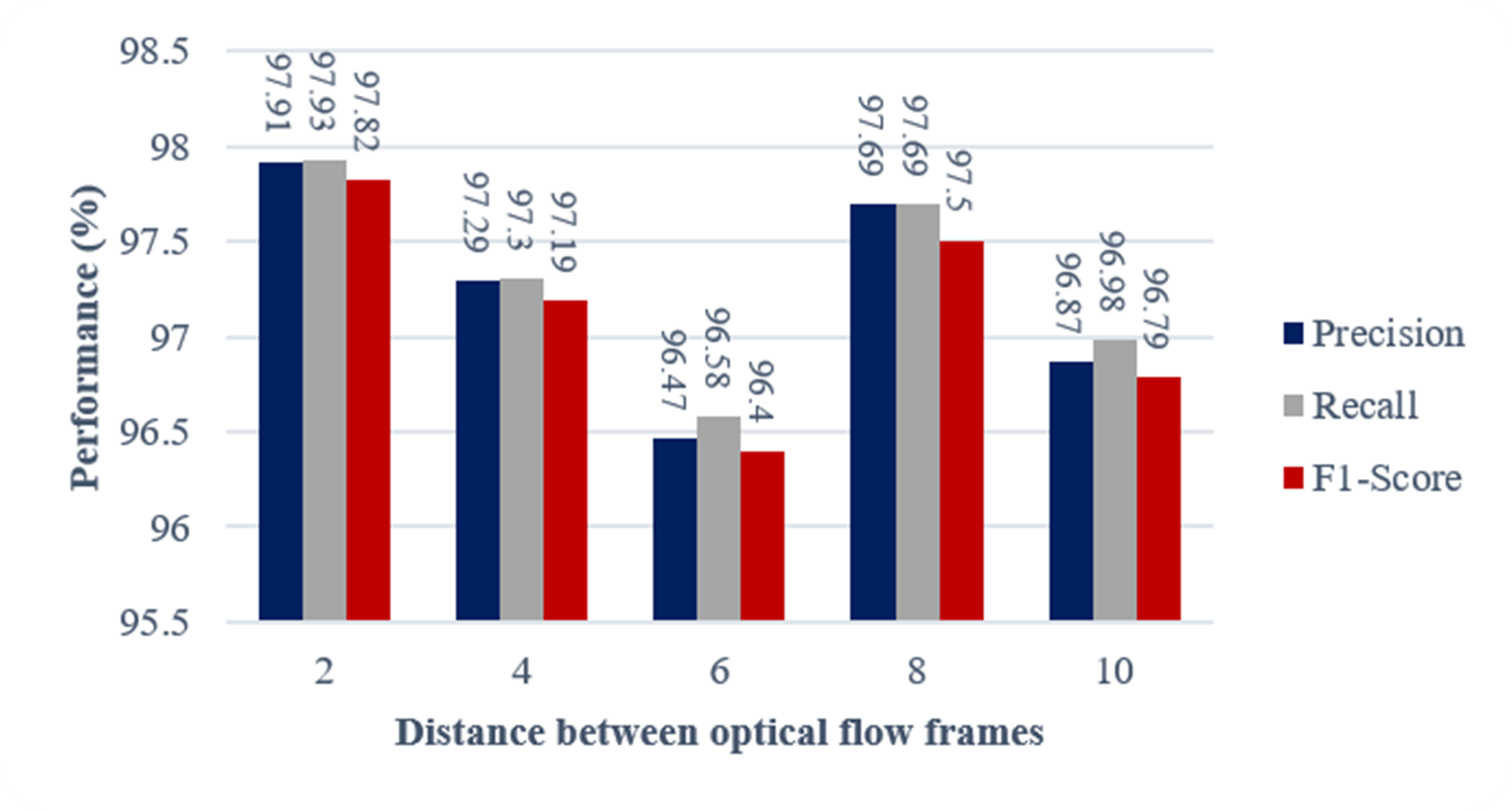}
        \caption{}
    \end{subfigure}%
    \\
    \begin{subfigure}[t]{\textwidth}
        \centering
        \includegraphics[height=1.7in]{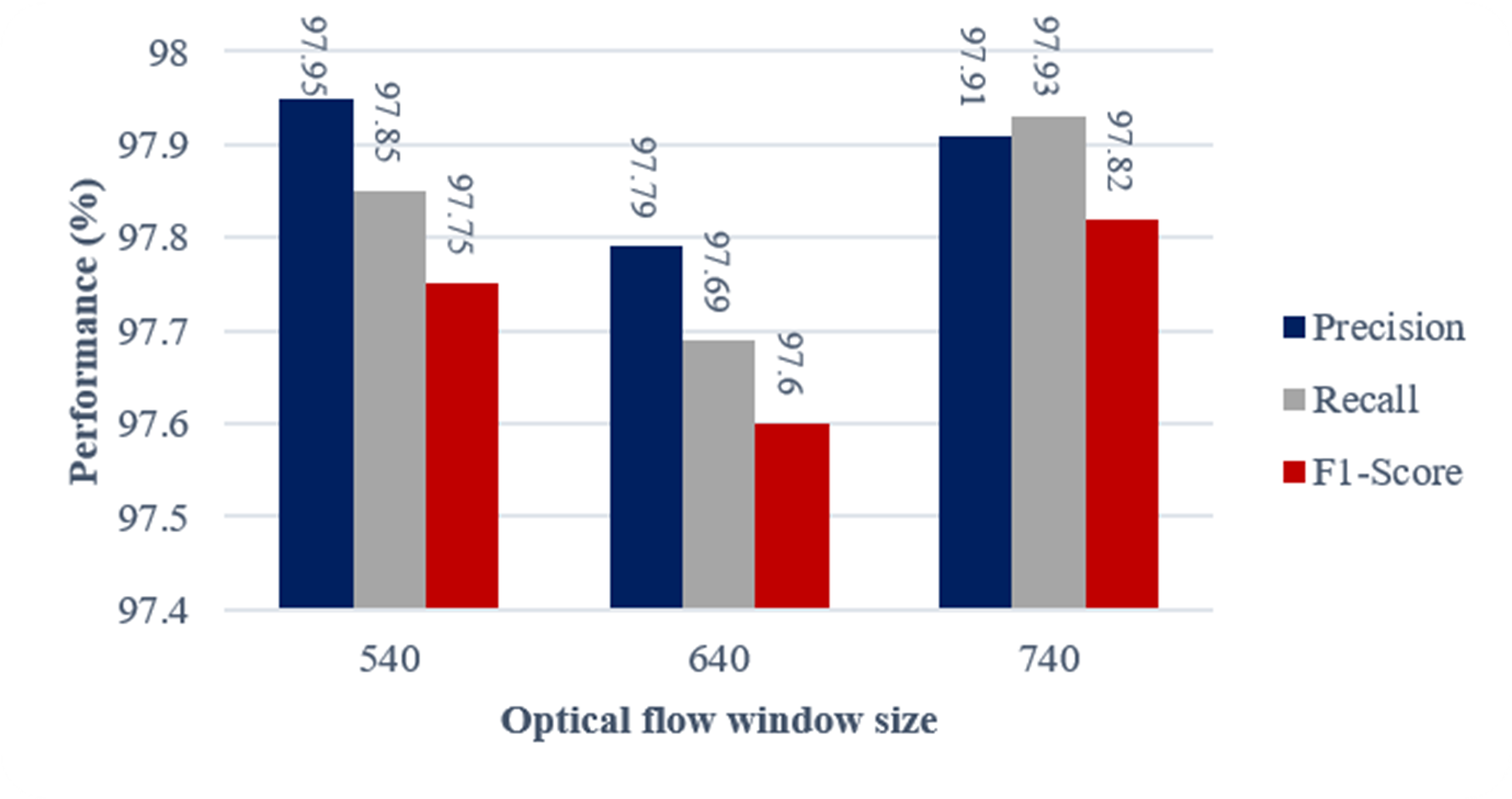}
        \caption{}
    \end{subfigure}
    \caption{(a) The algorithm performance in the KTH dataset for different values of the previous frames' color intensity, (b) The algorithm performance for calculating optical flow at various distances between frames, (c) The algorithm performance for different optical flow window sizes.}
    \label{fig:kthexp}
\end{figure*}

KTH data collection provided by Schuldt et al. \cite{Schuldt2004} includes six categories of human actions: Boxing, Hand Clapping, Hand Waving, Jogging, Running, and walking with a static background. Fig. \ref{fig:kthexp} represents the test results for different values of the color intensity of the frames, the distance between the optical flow frames, and the optical flow window size.

As shown in Table \ref{table:kthcomparison}, the proposed method achieves better recognition accuracy than the other approaches on the KTH dataset.
\begin{table}[]
    \centering
    \renewcommand{\arraystretch}{1.35}
    \caption{Comparison of our approach results with other methods on the KTH dataset}
    \label{table:kthcomparison}
    \begin{tabular}{  c  c  c  } 
    \noalign{\hrule height 1.5pt}
    \textbf{} & \textbf{Method} & \textbf{Accuracy $(\%)$} \\ 
    \noalign{\hrule height 1.5pt}
    Singh et al. \cite{Singh2019} & Histogram equalization enhancement & 94.5
    \\
    Abdelbaky et al. \cite{Abdelbaky2020} & PCANet-1 & 85.5
    \\
    Abdelbaky et al. \cite{Abdelbaky2020} & PCANet-2 & 90.47
    \\
    Chou et al. \cite{Chou2018} & Nearest Neighbor Classifier (NNC) & 89.31
    \\
    Chou et al. \cite{Chou2018} & Gaussian Mixture Model Classifier (GMMC) & 90.21
    \\
    Chou et al. \cite{Chou2018} & Nearest Mean Classifier (NMC) & 90.58
    \\
    Vishwakarma et al. \cite{Vishwakarma2020} & Gabor Ridget Transform & 96.66
    \\
    \textbf{Ours} & \textbf{Velocity and Direction Features} & \textbf{97.95}
    \\
    \noalign{\hrule height 1.5pt}

\end{tabular}
\end{table}

\subsection{Weizmann Dataset}
\begin{figure*}[t!]
    \centering
    \begin{subfigure}[t]{\textwidth}
        \centering
        \includegraphics[height=1.7in]{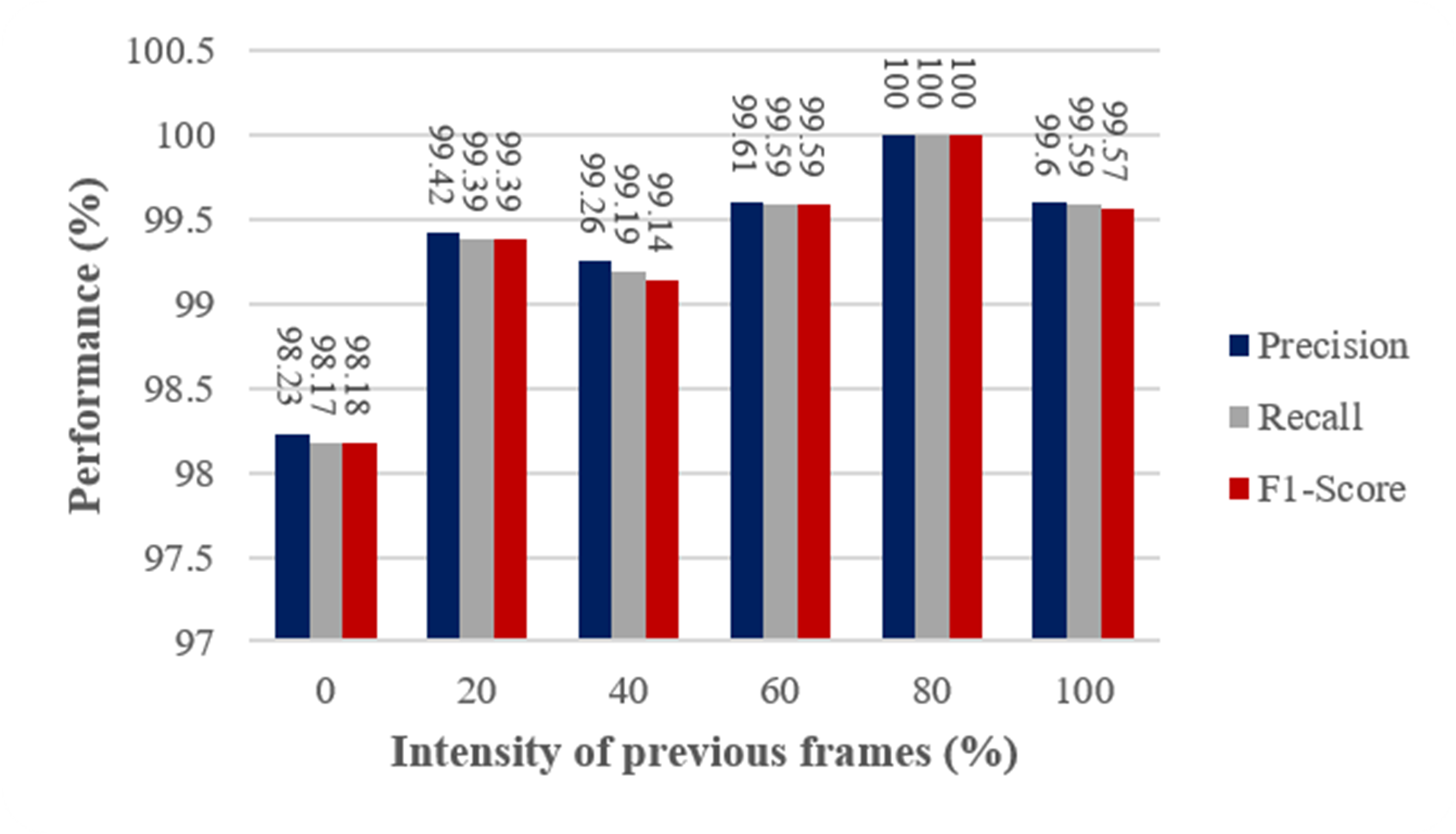}
        \caption{}
    \end{subfigure}%
    \\
    \begin{subfigure}[t]{\textwidth}
        \centering
        \includegraphics[height=1.7in]{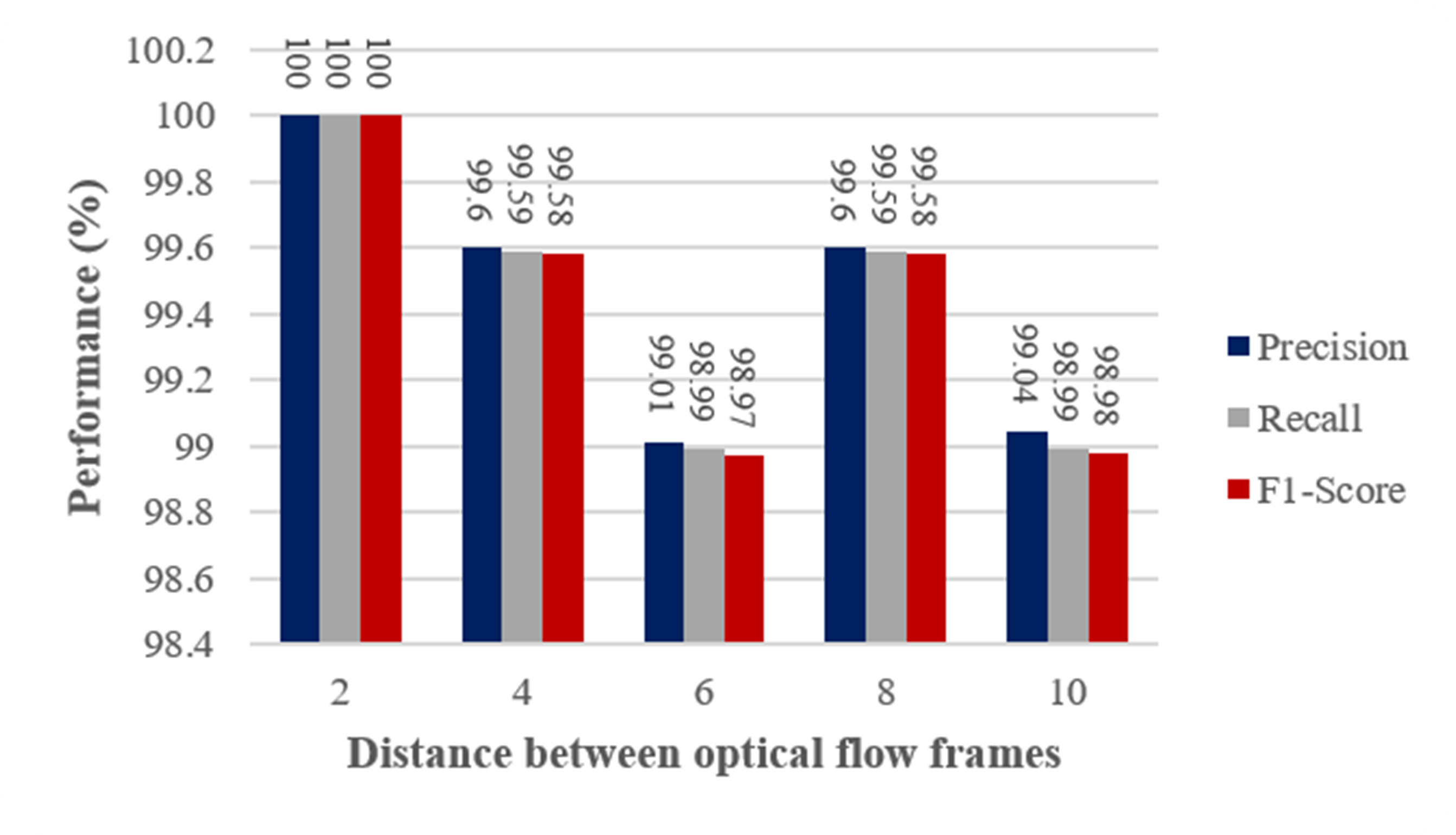}
        \caption{}
    \end{subfigure}%
    \\
    \begin{subfigure}[t]{\textwidth}
        \centering
        \includegraphics[height=1.7in]{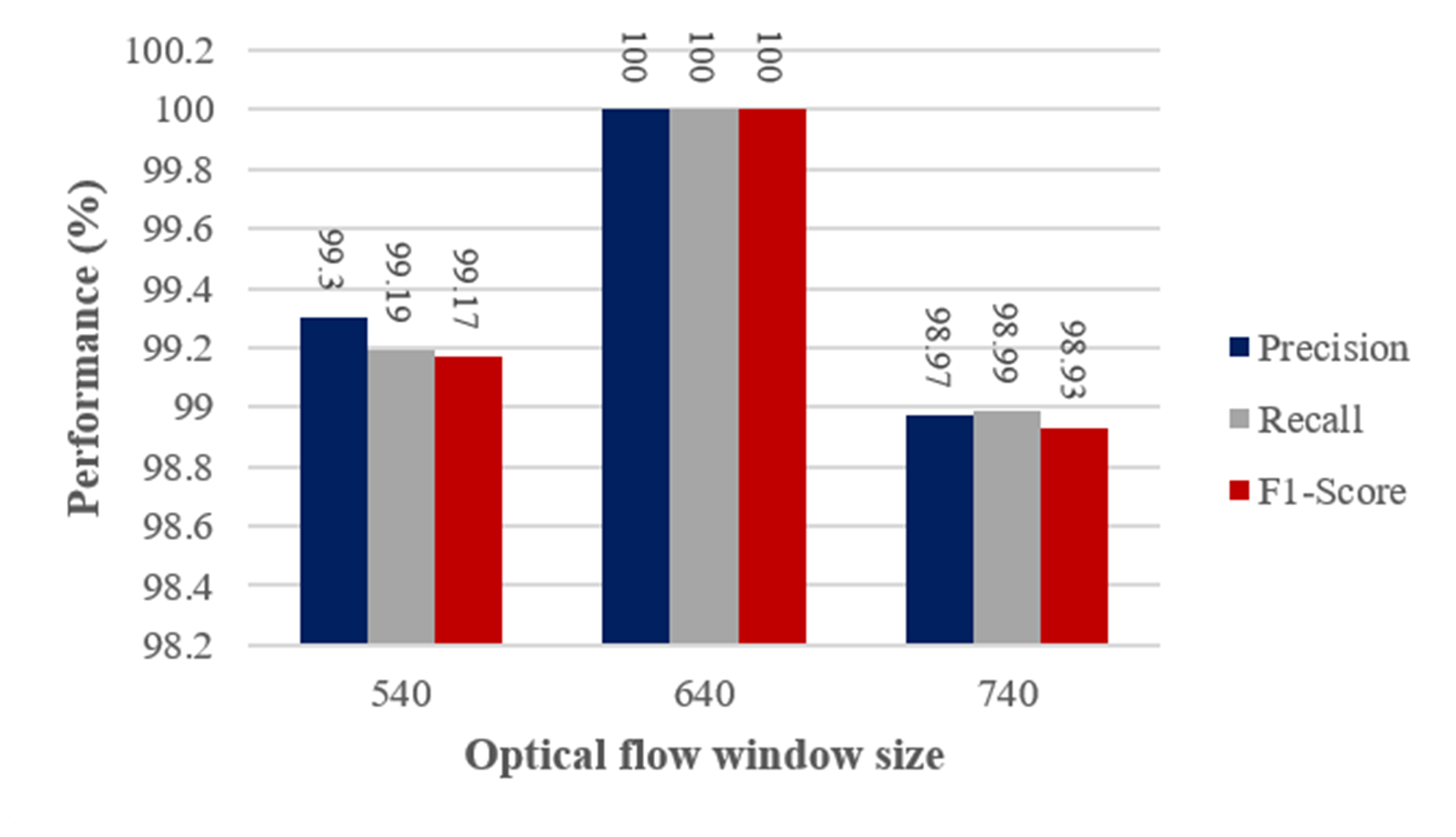}
        \caption{}
    \end{subfigure}
    \caption{(a) The algorithm performance in the Weizmann dataset for different values of the previous frames' color intensity, (b) The algorithm performance for calculating optical flow at various distances between frames, (c) The algorithm performance for different optical flow window sizes.}
    \label{fig:weizmannexp}
\end{figure*}

Weizmann data collection consists of 10 basic actions with a static background developed by the Weizmann Institute of Science \cite{Blank2005}. The background of the data set is simple, and the front-facing camera receives input from only one person. 
Fig. \ref{fig:weizmannexp} depicts the experiment results for different values of the color intensity of the frames, the distance between the optical flow frames, and the optical flow window size.

As shown in Table \ref{table:weizmanncomparison}, the proposed method achieves better recognition accuracy than the other approaches on the Weizmann dataset.
\begin{table}[]
    \centering
    \renewcommand{\arraystretch}{1.35}
    \caption{Comparison of our approach results with other methods on the Weizmann dataset}
    \label{table:weizmanncomparison}
    \begin{tabular}{  c  c  c  } 
    \noalign{\hrule height 1.5pt}
    \textbf{} & \textbf{Method} & \textbf{Accuracy $(\%)$} \\ 
    \noalign{\hrule height 1.5pt}
    Singh et al. \cite{Singh2019} & Histogram equalization enhancement & 97.66
    \\
    Abdelbaky et al. \cite{Abdelbaky2020} & PCANet-1 & 97.8
    \\
    Abdelbaky et al. \cite{Abdelbaky2020} & PCANet-2 & 100
    \\
    Chou et al. \cite{Chou2018} & Nearest Neighbor Classifier (NNC) & 87.78
    \\
    Chou et al. \cite{Chou2018} & Gaussian Mixture Model Classifier (GMMC) & 91.11
    \\
    Chou et al. \cite{Chou2018} & Nearest Mean Classifier (NMC) & 95.56
    \\
    Vishwakarma et al. \cite{Vishwakarma2020} & Gabor Ridget Transform & 96
    \\
    \textbf{Ours} & \textbf{Velocity and Direction Features} & \textbf{100}
    \\
    \noalign{\hrule height 1.5pt}

\end{tabular}
\end{table}

\subsection{HMDB-51 Dataset}
\begin{figure*}[t!]
    \centering
    \begin{subfigure}[t]{\textwidth}
        \centering
        \includegraphics[height=1.7in]{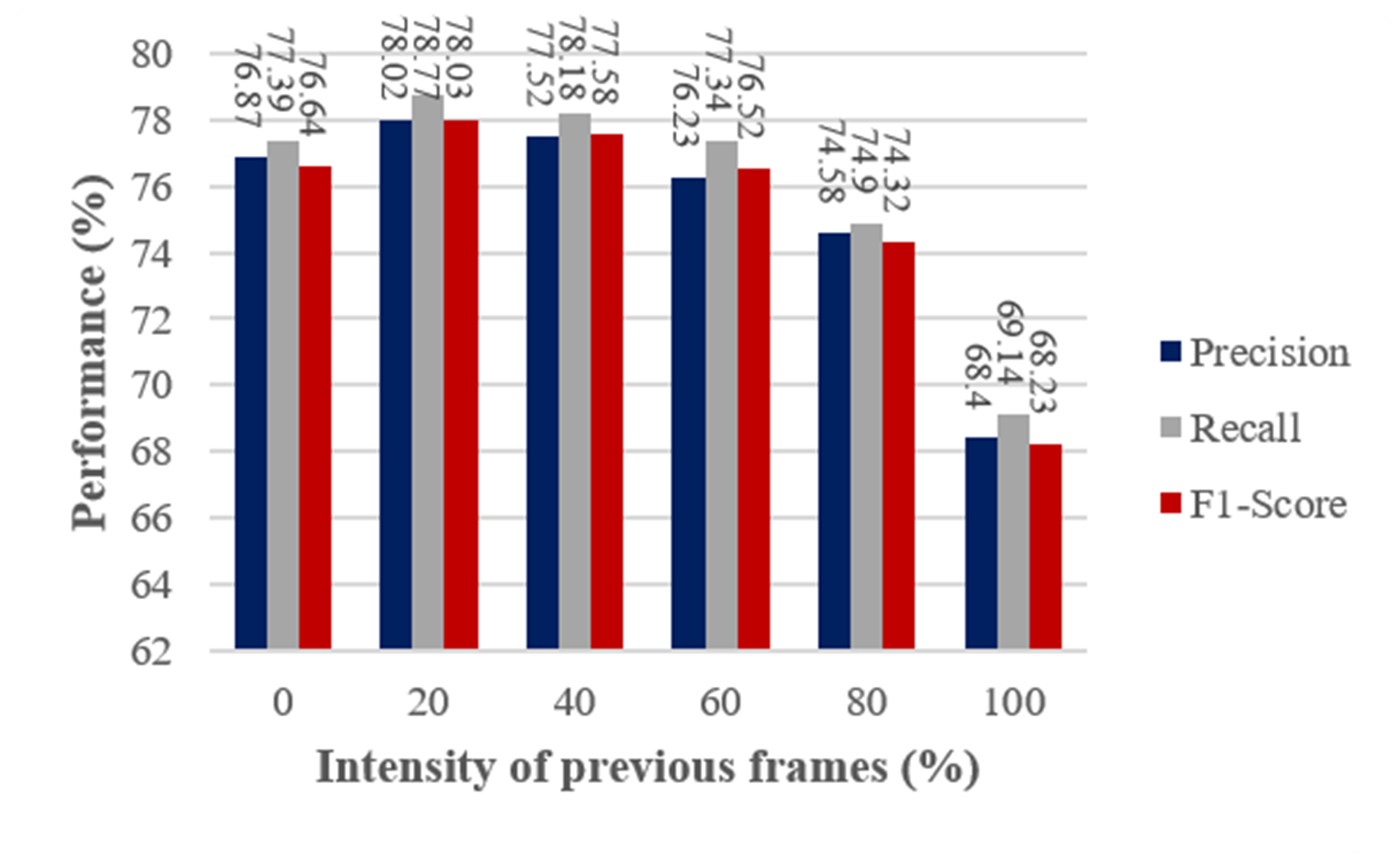}
        \caption{}
    \end{subfigure}%
    \\
    \begin{subfigure}[t]{\textwidth}
        \centering
        \includegraphics[height=1.7in]{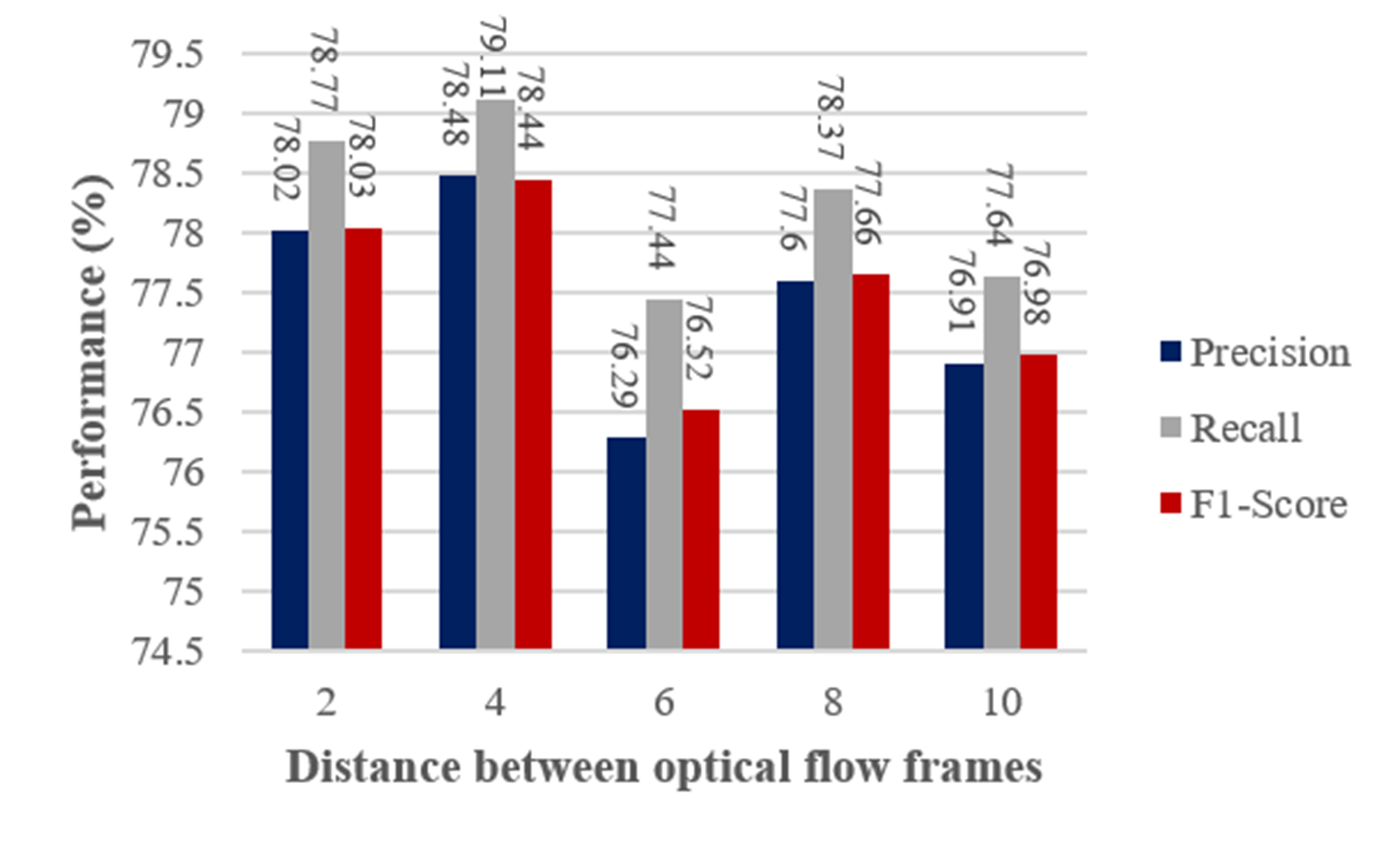}
        \caption{}
    \end{subfigure}%
    \\
    \begin{subfigure}[t]{\textwidth}
        \centering
        \includegraphics[height=1.7in]{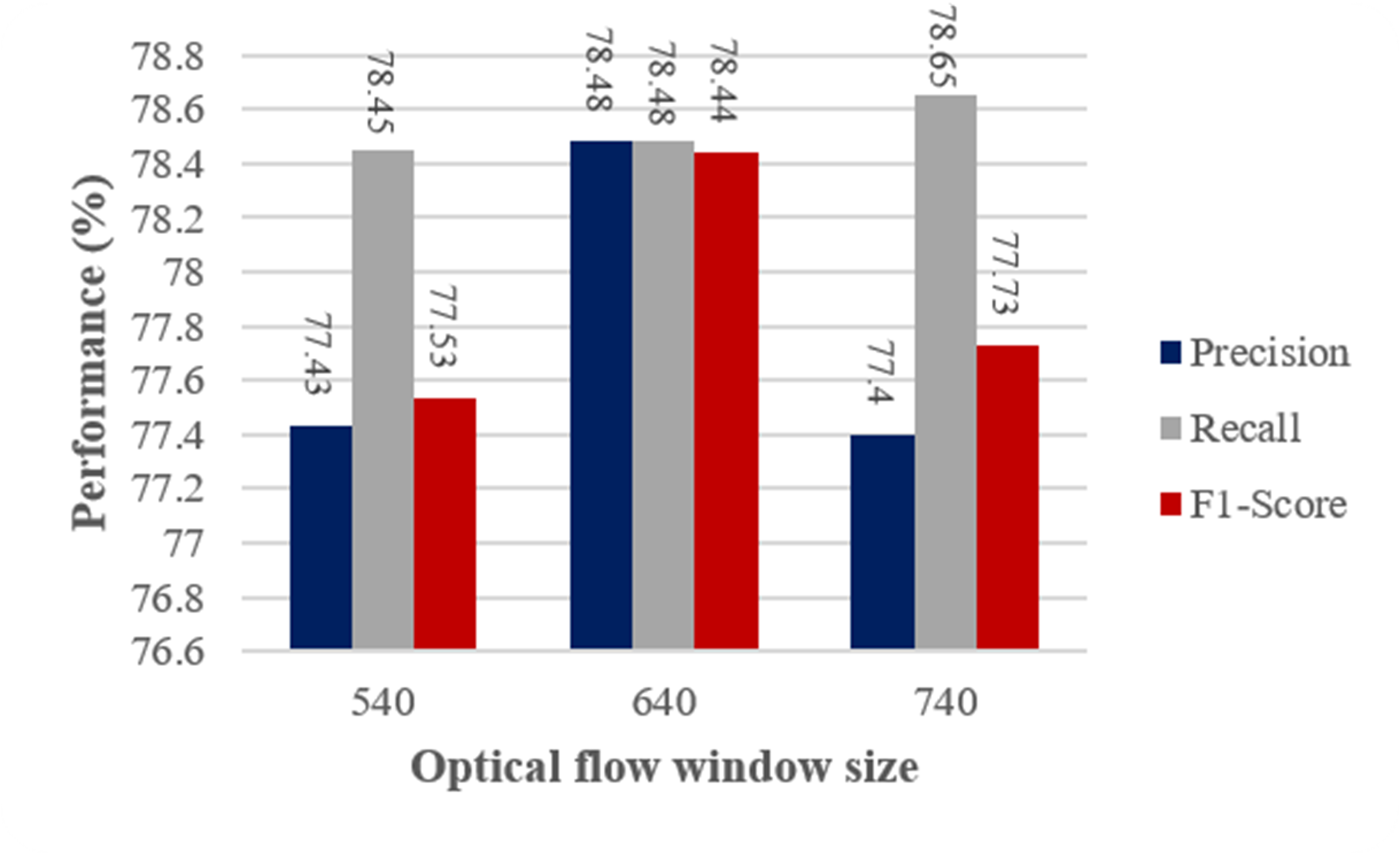}
        \caption{}
    \end{subfigure}
    \caption{(a) The algorithm performance in the HMDB-51 dataset for different values of the previous frames' color intensity, (b) The algorithm performance for calculating optical flow at various distances between frames, (c) The algorithm performance for different optical flow window sizes.}
    \label{fig:hmdb51exp}
\end{figure*}

The HMDB-51 dataset is one of the largest available datasets for activity recognition created in 2011 by the Serre lab at Brown University in the United States \cite{Kuehne2011}. It consists of 51 different types of everyday life acts from multiple outlets such as movies, YouTube, and Google videos. Since it includes scenes with more complex backgrounds, this dataset is more challenging than other datasets.
The test results for different values of frame color intensity, the distance between optical flow frames, and optical flow window size are shown in Fig. \ref{fig:hmdb51exp}.

Table \ref{table:hmdb51comparison} compares the results of our approach and other methods on the HMDB-51 dataset to assess the performance of this method.
\begin{table}[]
    \centering
    \renewcommand{\arraystretch}{1.35}
    \caption{Comparison of our approach results with other methods on the HMDB-51 dataset}
    \label{table:hmdb51comparison}
    \begin{tabular}{  c  c  c  } 
    \noalign{\hrule height 1.5pt}
    \textbf{} & \textbf{Method} & \textbf{Accuracy $(\%)$} \\ 
    \noalign{\hrule height 1.5pt}
    Asghari-Esfeden \cite{Asghari-Esfeden2020} & Dynamic Motion Representation & 84.2
    \\
    Zhang et.al \cite{Zhang2018} & Deeply-Transferred Motion Vector & 53
    \\
    Sun et al. \cite{Sun2017} & Lattice LSTM ($L^2$STM) & 66.2
    \\
    Ullah et al. \cite{Ullah2017} & DB-LSTM & 87.64
    \\
    Kar et al. \cite{Kar2017} & Adaptive Scan Pooling & 66.9
    \\
    \textbf{Ours} & \textbf{Velocity and Direction Features} & \textbf{78.48}
    \\
    \noalign{\hrule height 1.5pt}

\end{tabular}
\end{table}

\subsection{Real-Time Intention Recognition Embedded System}
Vivado design tools are used to synthesize and implement the system on the FPGA board and provide reports of power consumption, hardware resources used, and hardware schedule.
The Vivado Power Consumption Report identifies the chip's internal power, the chip's power budget, connection temperature, and the heat that the chip transmits to the environment per Watt ($\theta$JA). The power consumption report for the ZCU102 board is given in Table \ref{table:power}.
\begin{table}[]
    \centering
    \renewcommand{\arraystretch}{1.35}
    \caption{Power consumption and Temperature report for the ZCU102 board}
    \label{table:power}
    \begin{tabular}{  c  c  c  c  } 
    \noalign{\hrule height 1.5pt}
    \textbf{Total On-Chip Power} & \textbf{Junction Temperature} & \textbf{Thermal Margin} & \textbf{Effective $\theta$JA} \\ 
    \noalign{\hrule height 1.5pt}
    7.326 W & 32.2 $\degree$C & 67.8 $\degree$C (67.6 W) & 1.5 $\degree$C/W \\
    \noalign{\hrule height 1.5pt}

\end{tabular}
\end{table}
The setup timing report for the recognition system for the 333 MHz clock on the ZCU102 board is shown in Table \ref{table:timing}. The positive value of WNS in this report indicates the success of the paths generated by the hardware. It shows the difference between the delay of the critical path and the clock period, and being positive means that the setup time constraint set for the clock period is met. Violations of time constraints are indicated by TNS, where zero means that all complete design scheduling requirements are met.
\begin{table}[]
    \centering
    \renewcommand{\arraystretch}{1.35}
    \caption{Set up timing report for the ZCU102 board}
    \label{table:timing}
    \begin{tabular}{  c  c  c  c  } 
    \noalign{\hrule height 1.5pt}
    \multirow{2}{7em}{\textbf{Worst Negative Slack (WNS)}} & \multirow{2}{7em}{\textbf{Total Negative Slack (TNS)}} & \multirow{2}{8em}{\textbf{Number of Failing Endpoints}} & \multirow{2}{7em}{\textbf{Total Number of Endpoints}} \\ 
    \\
    \noalign{\hrule height 1.5pt}
     13.326 ns & 0 ns & 0 & 936503 \\
    \noalign{\hrule height 1.5pt}

\end{tabular}
\end{table}
In the holding delay report, the positive value of WHS indicates the success of the routes generated in terms of hold time, and its value indicates the minimum delay of the route to the flip flops. The total number of holding time violations with THS is shown to be zero, which means that all minimum hold time constraints in the design are met. The report on the minimum latency of the routes for the hold time analysis on the ZCU102 board is shown in Table \ref{table:holding_delay}.
\begin{table}[]
    \centering
    \renewcommand{\arraystretch}{1.35}
    \caption{Holding delay report for the ZCU102 board}
    \label{table:holding_delay}
    \begin{tabular}{  c  c  } 
    \noalign{\hrule height 1.5pt}
    \textbf{Worst Hold Slack (WHS)} & \textbf{Total Hold Slack (THS)} \\ 
    \noalign{\hrule height 1.5pt}
     0.009 ns & 0 ns\\
    \noalign{\hrule height 1.5pt}

\end{tabular}
\end{table}
The report in Table \ref{table:resource} shows the hardware resources used in the ZCU102 board.
\begin{table}[]
    \centering
    \renewcommand{\arraystretch}{1.35}
    \caption{Utilization and available hardware sources for the ZCU102 board}
    \label{table:resource}
    \begin{tabular}{  c  c  c  c  } 
    \noalign{\hrule height 1.5pt}
    \textbf{Resource} & \textbf{Utilization} & \textbf{Available} & \textbf{Utilization (\%)} \\ 
    \noalign{\hrule height 1.5pt}
    \textbf{LUT} & 126980 & 274080 & 46.33 \\
    \textbf{LUTRAM} & 17515 & 144000 & 12.16 \\
    \textbf{FF} & 277155 & 548160 & 50.56 \\
    \textbf{DSP} & 2070 & 2520 & 82.14 \\
    \textbf{BUFG} & 7 & 404 & 1.73 \\
    \noalign{\hrule height 1.5pt}

\end{tabular}
\end{table}
As we have said before, to ensure the real-time performance of the system, the processing rate is compared to the input rate. The entire system has a processing capacity of approximately 120 fps. In considering the fact that the standard USB camera receives 30 fps of data, we can be assured that our system operates in real-time. 

\section{Conclusion}
\label{sec:conclusion}
This article aims to save energy by improving automatic doors' efficiency by implementing a real-time embedded system for intention recognition. In this system, we used real-time activity detection to detect intent. To do this, by processing video data, we analyzed features such as the speed and direction of people's movements and developed a novel representation of video data. The crowded scene and the position of the camera do not affect the performance of our system. By modeling and using suitable processors for the embedded system, we have developed our system with low cost and high performance. At the beginning of this study, we collected the appropriate data, and after creating a representation of the video data, we used AlexNet CNN for classification, and we achieved 98.78\% recognition accuracy. To evaluate the efficiency of our data representation method, we also tested it on other datasets such as HMDB-51, KTH, and Weizmann, and we achieved an accuracy of 78.48\%, 97.95\%, and 100\%, respectively.
Finally, after creating the model, we simulated it on the ZCU102 board using Xilinx design tools, analyzed the power consumption and system timing, and simulated our system with a processing capacity of 120 fps on the ZCU102 board.
As mentioned earlier, despite the many methods used to implement activity recognition algorithms, this problem faces many challenges, and suggestions can be made to overcome them to continue this study. This algorithm can be used in human-computer or human-robot interfaces by preserving more detail and using gesture recognition.

\bibliographystyle{unsrt}  
\bibliography{references}

\end{document}